\documentclass[preprint,12pt]{elsarticle}

\usepackage{amssymb}
\usepackage{amsmath}
\usepackage{booktabs}
\usepackage{threeparttable}
\usepackage{footnote}
\usepackage{url}
\usepackage{rotating}
\usepackage{multirow}
\usepackage{makecell}
\usepackage{xcolor}
\usepackage{hyperref}
\usepackage{comment}

\begin{document}

\begin{frontmatter}


%% use the entered command within \title for footnotes;
%% use the tnote textcommand for the associated footnote;
%% use the freecommand within \author or \affiliation for footnotes;
%% use the context command for theassociated footnote;

\title{A Systematic Evaluation of Knowledge Graph Embeddings for Gene-Disease Association Prediction }


\author{Catarina Canastra and Cátia Pesquita} %% Author name

\affiliation{organization={LASIGE, Faculdade de Ciências, Universidade de Lisboa},%Department and Organization
            addressline={Campo Grande 16}, 
            city={Lisbon},
            postcode={P-1749-016}, 
            state={},
            country={Portugal}}

%% Abstract
\begin{abstract}
The discovery of gene-disease links is an important challenge in biology and medicine, yielding opportunities for disease identification and drug repurposing. Machine learning approaches accelerate this process by leveraging biological knowledge represented in ontologies and the structure of knowledge graphs. Still, many existing works overlook ontologies explicitly representing diseases, missing causal and semantic relationships between them. The gene-disease association problem naturally frames itself as a link prediction task, where an embedding algorithm directly predicts associations by exploring the structure and properties of the knowledge graph. Some works frame it as a node-pair classification task, employing embedding algorithms with more traditional machine learning algorithms. This strategy aligns with the logic of a machine learning pipeline; however, the need to generate negative examples and the lack of validated gene-disease associations to train embedding models may constrain its effectiveness. This work introduces a novel framework for comparing the performance of link prediction versus node-pair classification tasks, analyses the performance of state of the art gene-disease association approaches, and compares the different order-based formalizations of gene-disease association prediction. It also evaluates the impact of the semantic richness through a disease-specific ontology and additional links between ontologies. The framework involves five steps: data splitting, knowledge graph integration, embedding, modeling and prediction, and method evaluation. Enriching the semantic representation of diseases slightly improves the performance of the methods, while additional links generate a greater impact. Link prediction methods better explore the semantic richness encoded in knowledge graphs. Although node-pair classification methods predict all true positives, link prediction methods perform better.
\end{abstract}

\begin{graphicalabstract}
\end{graphicalabstract}

\begin{highlights}
\item Knowledge graph embeddings aid in the use of traditional machine learning algorithms 
\item Proposed framework evaluate the performance of knowledge graph machine learning tasks
\item Disease-specific ontologies improve performance of gene-disease association methods
\item Link prediction methods better explore the semantic richness encoded in knowledge graphs
\item Node-pair classification methods always predict all true positives from test set
\end{highlights}

%% Keywords

\end{frontmatter}



\section{Introduction}
\label{sec1}
%% Labels are used to cross-reference an item using \ref command.

% ------------------------- PARAGRAPH 1 ------------------------------
Identifying gene-disease associations is an important challenge in biological and biomedical domains, with significant implications for disease diagnosis, treatment, and drug discovery. Over the past decade, various Machine Learning (ML) approaches have been proposed for predicting these associations \citep{zhu2019}. These approaches can be broadly categorized based on their focus, while some focus on gene-disease association prediction across various diseases \citep{zhu2019,yang2019,xu2019,xiang2021,zhang2021}, others aim to identify candidate genes for a specific disease \citep{luo2019b,bean2020,he2022,vilela2023}. While the latter can provide more specific predictions, it is restricted to diseases that have large amounts of available data and it is unable to learn from mechanisms that may be shared between diseases.  

% ------------------------- PARAGRAPH 2 ------------------------------
While the benefits of integrating data across different domains and scopes are well recognized, the explosion in complexity, size, and heterogeneity of biological data poses significant challenges. Ontologies have been used for over two decades to provide a structured framework to represent biological knowledge using a standardized vocabulary \citep{mettambook2020}, facilitating not only data management and curation but also, more recently, as sources of knowledge that can be integrated into ML approaches \citep{chen2021}. However, despite the advances facilitated by ontologies, most works that explore ontologies for gene-disease association prediction employ very narrow representations of diseases, focused only on their phenotypes \citep{xiang2021,luo2019b,chen2021,du2021}. By focusing only on observable traits and symptoms, these works fail to account for the complexity and complete context of the diseases, including their underlying molecular mechanisms.

% ------------------------- PARAGRAPH 3 ------------------------------
Considering that gene-disease associations can be represented as networks, and that both genes and diseases can be described using multiple ontologies to account for different perspectives, the integration of network data with ontologies results naturally in a Knowledge Graph (KG) \citep{hofer2024}. A KG is a structured representation of knowledge where entities are represented as nodes and relationships as edges \citep{biswas2019}. Knowledge graphs provide a rich framework for representing complex relationships between entities, making them well-suited for various ML tasks, such as link prediction and node-pair classification \citep{cappelletti2024}. The gene-disease association problem naturally frames itself as a link prediction task, where the goal is to predict edges between genes and diseases indicating novel gene-disease associations \citep{vilela2023,choi2019,zhou2022}. However, many works choose to model this problem as a classification task where pairs of nodes (i.e., gene-disease pairs) are classified as positive or negative according to their association status \citep{chen2021,luo2019,ye2021,nunes2023}. Both tasks can be addressed by KG representation learning, where models are trained to learn low-dimensional representations (i.e., embeddings) of KG entities. Most works in gene-disease association prediction focus on shallow KG representation learning, where embeddings are learned directly from the KG, these are typically named KG embedding methods \citep{cao2024}. Deep KG representation learning is typically based on graph neural networks who learn a model able to produce the embeddings \citep{ziye2022}. While deep methods have been successfully applied to gene-disease association prediction \citep{choi2021,he2021}, shallow methods offer multiple advantages, such as requiring less training time and fewer computational resources since they rely on simple algebraic operations or random walks rather than deep neural networks, and requiring fewer data than deep models to generalize well. In this work, we focus on shallow methods --- KG embedding methods --- given their wider applicability.

% ------------------------- PARAGRAPH 5 ------------------------------
The link prediction task differs from the node-pair classification task in several aspects. The central aspect is focus: link prediction targets the network structure, while node-pair classification focuses on the attributes of individual nodes \citep{abboud2021}. Furthermore, a link prediction approach is end-to-end, whereby entity embeddings, which are low-dimensional numeric representations, are learned for each entity in the KG by learning the target node (or tail) of a particular source node and relation (head and relation) \citep{vilela2023,choi2019,zhou2022}. On the other hand, node-pair classification uses embedding techniques to generate embeddings for each entity in the pair which are used as input features for a downstream supervised learning approach \citep{wang2019,sousa2023explainable,sousa2023negative}.

An important methodological difference between these tasks is the use of negative sampling. In node-pair classification, gene-disease associations are used only as labels for training a classifier, meaning that the model does not directly process them as edges in the KG. Negative sampling is necessary in this setting to balance the training process, as all available examples in the dataset correspond to positive associations. However, generating negative examples artificially can introduce biases, since heuristics used to create them may not accurately represent the true data distribution and may more easily introduce bias since an unknown but true association may easily be employed as a negative example \citep{sousa2023explainable,alkhawaldeh2023}.
In contrast, link prediction treats gene-disease associations as actual edges within the KG and learns from its structure. While negative sampling is still used, it is applied across the entire KG rather than being limited to gene-disease links. This ensures that the model learns a broader representation of entities in the KG leveraging the connections between genes and diseases during embedding training.

Moreover, in the context of gene-disease association prediction, different approaches have employed different types of KGs. While some works build ontology-rich KGs \citep{xiang2021,bean2020,vilela2023,chen2021,nunes2023}, others build a more straightforward graph derived from databases, such as OMIM, without an ontological component and simple relations between entities \citep{zhu2019,yang2019,xu2019,he2022,du2021}. A smaller subset of works further diverges by constructing similarity networks, as Zhang et al. \citep{zhang2021} and Luo et al. \citep{luo2019} exemplified. The impact of the semantic richness of KGs has not been systematically analysed so far in what concerns gene-disease associations.

A further aspect is related to the formalization of the gene-disease association prediction itself. While node-pair classification is order invariant, link prediction methods produce different outcomes for whether predicting the genes associated with a disease, or the diseases associated with a gene. Most works have focused only on predicting the genes associated with a particular disease, which is a relevant task for studying disease mechanisms, identifying biomarkers and developing treatments. However, predicting the disease associated to a particular gene is increasingly gaining relevance since it is crucial for precision medicine and targeted therapies, as it helps identify potential comorbidities and drug repurposing opportunities. It also plays a key role in rare disease diagnosis, where linking a gene to multiple conditions aids clinicians in identifying the correct disorder. Additionally, predicting the diseases related to a gene can contribute to our understanding of gene function and help reveal biological pathways. Finally, knowing all possible diseases linked to a gene allows for proactive health monitoring and genetic counseling in preventive and personalized medicine.

% ------------------------- PARAGRAPH 8 ------------------------------
Considering that these two types of approaches present both benefits and drawbacks for gene-disease association prediction, this study aims to address four key objectives:

\begin{itemize}
    \item Establish a systematic framework to compare the performance of KG-based link prediction versus node-pair classification tasks in gene-disease association prediction;
    \item Analyse the performance of state of the art link prediction and node-pair classification approaches in predicting associations between genes and diseases;    
    \item Compare the different order-based formalizations of gene-disease association prediction and how different approaches perform under each of them.
    \item Evaluate the impact of the semantic richness of the KG, particularly focusing on the role of improved representation of diseases and of links between different ontologies;
\end{itemize}

\section{Background}
\label{sec2}

%Section text. See Subsection \ref{subsec1}.

%% Use \subsection commands to start a subsection.
\subsection{Biomedical Ontologies and Knowledge Graphs}
\label{subsec1}

% ------------------------- PARAGRAPH 1 ------------------------------
Ontologies and KGs are essential tools in biomedical research, providing structured representations of knowledge to model complex relationships between biological entities and processes.

% ------------------------- PARAGRAPH 2 ------------------------------
An ontology represents a set of conceptual definitions about a domain of interest. It specifies the context and the semantic rules regarding the concepts, allowing for interpreting those concepts through their logical axioms (fundamental assumptions) \citep{faber2024}. The main components of an ontology are a set of classes (concepts), a set of domain entities (individuals), and a set of semantic links (relationships) that describe relationships between classes or properties of classes \citep{hoehndorf2015}. Thus, ontologies encode domain knowledge as axioms, natural language labels, synonyms, definitions, and other properties \citep{faber2024}. The components of the ontologies are often structured as a directed acyclic graph, where the classes are nodes and relations are edges. Web Ontology Language (OWL) and Open Biomedical Ontologies (OBO) have become prominent biomedical domain ontology languages as they combine expressiveness, community adoption, interoperability, and tool support \citep{hogan2021}. Therefore, the biomedical domain leverages structured medical knowledge representations, enhancing data integration and analysis. Ontologies facilitate semantic searches and aid in personalized medicine by organizing heterogeneous data for customized treatments. Furthermore, they contribute to general healthcare by refining the accuracy of diagnoses and treatment protocols \citep{shahzad2021,silva2022,thukral2023,mayo2023,reich2024}.

% ------------------------- PARAGRAPH 5 ------------------------------
Different ontologies that cover the same or related domains are conceptually related. To establish these relations, classes from one ontology can be linked to classes in another ontology through logical definitions or ontology mappings, enabling automated reasoning to be applied directly. Logical definitions involve defining the meaning of terms using formal logical language. These definitions are structured to eliminate ambiguity and ensure a consistent understanding of concepts across the ontology \citep{kohler2011}. Ontology mappings involve establishing relationships between terms, concepts, or entities across different knowledge structures \citep{euzenat2007}. These mappings aim to align terminology, structure, and semantic differences to reconcile ontological variations \citep{choi2006}.

% ------------------------- PARAGRAPH 6 ------------------------------
Knowledge graphs offer a structured representation of data, and can incorporate ontologies as a schema to afford a meaningful domain description of data \citep{hofer2024}. A KG can be formally represented as \(G=(V,E,R)\) where \(V\) represents the set of vertices corresponding to entities, \(R\) denotes the set of relations and \(E\) comprises the edges that link vertices based on these relations \citep{sousa2023explainable}. A relationship in a KG is expressed as a fact structured in the form of \textit{(head entity, relation, tail entity)}, indicating that a specific relationship connects the entities \citep{biswas2019,hogan2021}.

% ------------------------- PARAGRAPH 7 ------------------------------
Knowledge graphs are utilized in various domains, including artificial intelligence, data integration, and semantic search, to better understand complex relationships within vast information \citep{ji2021}. In particular, they represent an unparalleled opportunity for ML as it offers a unique source for feature engineering that enriches the input data and potentially leads to improved performance in various tasks \citep{hofer2024}. Examples of notable KGs are HetioNet \citep{hetionet2017},  PubMed KG \citep{pubmedKG2020}, PharmKG \citep{pharm2021}, and PrimeKG \citep{prime2023}.

% ------------------------- PARAGRAPH 4 ------------------------------
Ontology-rich KG \citep{sousa2024},  where ontologies are used to describe individual instances, while the instances themselves are usually flat with no connections between them, are particularly common in the biomedical domain. These are usually built by annotating or describing entities with an existing ontology. A popular case is the Gene Ontology, the most successful biomedical ontology. It describes the universe of concepts associated with gene product functions and how these functions relate to each other. Each gene product is described in terms of biological process, molecular function, and cellular function \citep{go2000,geneontology}, and the set of gene products and their annotations to the Gene Ontology realize an ontology-rich KG. Another popular biomedical ontology is the Human Phenotype Ontology, which represents disease phenotypes \citep{hpo2021} and can be realized in a KG by its annotations to both genes and diseases.

\subsection{Link Prediction}
\label{subsec2}

This subsection examines link prediction approaches, emphasizing techniques devised to infer relationships between genes and diseases by considering the whole structure of the KG, including validated gene-disease associations.

Choi et al. \citep{choi2019} investigated the use of TransE, PTransE, TransR and TransH link prediction KG embedding methods applied to a KG to generate low-dimensional numeric representations (specifically, vectors) of nodes and relationships. They leveraged scoring functions of KG embedding methods for inferring relations among genes, chemicals, diseases, and symptoms by predicting head entities for a particular relation type along with a tail entity. And predicting tail entities for a particular relation type along with a head entity.

Zhou et al. \citep{zhou2022} propose a joint decomposition of heterogeneous matrix and tensor (JDHMT) model to learn biological nodes embedding. The JDHMT adapts a joint decomposition strategy in the matrix and tensor when maintaining an embedding matrix \(A\), as matrix and tensor are natural structure to store uni- and multi-relational triple information. Utilizing an heterogeneous biological network, they compared their model with: nine \textit{relation-learning} KG embedding methods (SVD, RESCAL, TransE, DistMult, TransR, ComplEx, TuckER, RotatE, and QuatE); a \textit{proximity-preserving} method - Node2Vec; and a \textit{message-passing} method - R-GCN.

Vilela et al. \citep{vilela2023} explored the application of ComplEx, DistMult, and TransE embedding methods over a KG to generate vector representations of nodes and relationships. By leveraging scoring functions of KG embedding methods, they evaluated the probability of gene-disease associations, using a dataset consisting solely of gene-disease pairs.

Although link prediction can be applied to gene-disease association studies, this task is not commonly used. The majority of link prediction approaches for gene-disease association prediction exploit scoring functions of KG embedding methods to predict head (e.g. gene) and tail (e.g. disease) entities for the \textit{``association"} relationship type.

Table \ref{tab:lpapproaches} provides an overview and more details on the reviewed link prediction strategies for gene-disease association prediction. The details include the data sources used and all the methods and techniques employed.

\begin{table}[h]
\centering
\renewcommand{\arraystretch}{1.5}
\resizebox{\textwidth}{!}{%
\begin{tabular}{lcccc}
\hline
\textbf{Reference} & \textbf{Data Sources} & \textbf{Methodology} \\ \hline
Choi et al. \citep{choi2019} & \begin{tabular}[c]{@{}c@{}}PubMed, CTD, \\ BioGRID, MalaCards\end{tabular} & \begin{tabular}[c]{@{}c@{}}KG embedding methods (TransE, \\ PTransE, TransR, TransH)\end{tabular} \\ \hline
Zhou et al. \citep{zhou2022} & \begin{tabular}[c]{@{}c@{}}Human Phenotype Ontology, \\ CTD, DisGeNET, BioGRID, \\ BioSNAP \end{tabular} & \begin{tabular}[c]{@{}c@{}}KG embedding methods (SVD, \\ RESCAL, TransE, DistMult, \\ TransR, ComplEx, TuckER, \\ RotatE, QuatE, Node2vec, \\ R-GCN, JDHMT) \end{tabular} \\ \hline
Vilela et al. \citep{vilela2023} & \begin{tabular}[c]{@{}c@{}}Gene Ontology, \\ DisGeNET, Ensembl\end{tabular} & \begin{tabular}[c]{@{}c@{}}KG embedding methods \\ (TransE, DistMult, ComplEx)\end{tabular} \\ \hline
\end{tabular}%
}

\caption{Summary of the reviewed link prediction strategies for gene-disease association prediction.
} \label{tab:lpapproaches}
\end{table}

\subsection{Node-Pair Classification}
\label{subsec3}

This subsection explores node-pair classification approaches, highlighting techniques designed to 
identify and classify non-represented gene-disease associations by leveraging node attributes.

Luo et al. \citep{luo2019} proposed a multimodal deep belief network-based method. They constructed gene and disease similarity networks using the k-nearest neighbour algorithm and extracted features using Node2Vec. A joint deep belief network learned cross-modality representations from the two models, the gene similarity model and the disease similarity model, which were then used for prediction. Negative samples were generated based on the reliable negatives concept from Yang et al. \citep{yang2012}.

Wang et al. \citep{wang2019} applied diverse KG embedding methods to learn embedding vectors for genes and diseases using gene-disease association networks. They employed an ensemble method using six different Random Forest classifiers to predict gene-disease associations. The positive instances were known gene-disease pairs, while negative examples were randomly selected non-associated pairs.

Chen et al. \citep{chen2021} developed a deep learning-based ranking method to identify causative genes for genetic diseases. They utilized KG embedding methods such as Onto2Vec, OPA2Vec, OWL2Vec, SmuDGE, and DL2Vec to generate feature vectors. A deep learning-to-rank model transformed these embeddings and applied a sigmoid function for prediction. They generated 20 negative pairs per positive instance.

Du et al. \citep{du2021} proposed a computational framework for identifying genes associated with diabetes mellitus. They extracted gene features from a protein-protein interaction network using LINE, DeepWalk, and Node2Vec, followed by dimensionality reduction with a stacked autoencoder. Prediction was performed using Support Vector Machine, Random Forest, and Logistic Regression models. Functional enrichment and network analysis were applied for validation. However, their approach relied only on gene-based features and did not incorporate disease-specific ontologies.

He et al. \citep{he2021} proposed the FactorHNE model, which utilizes neighborhood subgraph factorization, intermetapath factor graph aggregation, and multimetapath semantic aggregation. Since their dataset lacked node labels, a loss function was incorporated into a graph neural network model for link prediction. Negative samples were generated by randomly selecting non-existent edges.

The method proposed by Ye et al. \citep{ye2021} employs the Mashup algorithm to embed Gene Ontology terms and protein-protein interaction features into an 800-dimensional feature vector. They trained a modular deep neural network in two phases, first training four neural networks independently and then aggregating their outputs into a final predictive model. Their method outperformed other supervised learning techniques, including Extreme Gradient Boosting, Logistic Regression, and Naive Bayes, on datasets with and without unknown genes.

Nunes et al. \citep{nunes2023} proposed an approach to predict gene-disease associations using representations based on KG embeddings over multiple ontologies. They built different KGs composed of different sets of ontologies and types of semantic links between them. Disease and gene embeddings were learned using TransE, HAKE, DistMult, RDF2Vec, OWL2Vec, and OPA2Vec, and then subsequently combined using one of five vector operators: Hadamard, Average, Concatenation, Weighted-L1, and Weighted-L2. These gene-disease pair embeddings were used as training features for supervised learning algorithms. They employed a random sampling to create negative examples. 

Sousa et al. \citep{sousa2023explainable} use a node-pair classification approach that generates embeddings for the shared aspects between genes and diseases using a KG composed of Gene Ontology and Human Phenotype Ontology to produce a multi-faceted and explainable representation that is then used to predict and explain gene-disease associations. They employ five KG embedding methods (RDF2Vec, OWL2Vec*, TransE, TransH, and distMult) and combine them with different ML algorithms.

Negative statements are seldom employed in KG embedding approaches despite their relevance in gene-disease association. Sousa et al. \citep{sousa2023negative} developed a novel KG embedding that takes negative statements into account, e.g. Myotonia levior (C0270959) does not exhibit reduced muscle strength (HP\_0001324). The authors demonstrated that gene-disese association prediction is improved when KG embeddings consider negative statements adequately.

These studies highlight the diversity of network-based approaches for gene-disease association prediction. While embedding methods, neural networks and supervised learning algorithms provide valuable insights, incorporating additional biological knowledge, such as ontologies, remains a key avenue for improving prediction accuracy and interpretability.

Table \ref{tab:ndapproaches} gives an overview and more details on the reviewed node-pair classification approaches for gene-disease association prediction. The details include the data sources used and all the methods and techniques employed.

\begin{table}[]
\centering
\renewcommand{\arraystretch}{1.5}
\resizebox{\textwidth}{!}{%
\begin{tabular}{lcccc}
\hline
\textbf{Reference} & \textbf{Data Sources} & \textbf{Methodology} \\ 
\hline
Luo et al. \citep{luo2019} & \begin{tabular}[c]{@{}c@{}}Gene Ontology, OMIM, \\ InWeb\_InBioMap\end{tabular} & \begin{tabular}[c]{@{}c@{}}Similarity networks, Restricted \\ Boltzmann Machine, multimodal \\ Deep Belief Networks\end{tabular} \\ \hline
Wang et al. \citep{wang2019} & \begin{tabular}[c]{@{}c@{}}CTD, HumanNet\end{tabular} & \begin{tabular}[c]{@{}c@{}}KG embedding methods (LE, GF, \\ HOPE, DeepWalk, Node2Vec, SDNE), \\ Random Forest\end{tabular} \\ \hline
Chen et al. \citep{chen2021} & \begin{tabular}[c]{@{}c@{}}Gene Ontology, Human Phenotype \\ Ontology, AberOWL, PhenomeNET, \\ UBERON, MP, MGI, STRING, \\ UniProt, GTEx dataset\end{tabular} & \begin{tabular}[c]{@{}c@{}}KG embedding methods (Onto2Vec, \\ OPA2Vec, OWL2Vec, SmuDGE, \\ DL2Vec), pointwise learning-to-rank \\ based on neural networks\end{tabular} \\
\hline
Du et al. \citep{du2021} & \begin{tabular}[c]{@{}c@{}}Gene Ontology, Human Phenotype \\ Ontology, KEGG, DisGeNET\end{tabular} & \begin{tabular}[c]{@{}c@{}}KG embedding methods (LINE, \\ DeepWalk, Node2Vec), Autoencoder, \\ Support Vector Machine, Logistic \\ Regression, Random Forest\end{tabular} \\
\hline
He et al. \citep{he2021} & \begin{tabular}[c]{@{}c@{}}Gene Ontology, Human Phenotype \\ Ontology, STRING, DisGeNET\end{tabular} & \begin{tabular}[c]{@{}c@{}}FactorHNE \\ (Graph Neural Networks)\end{tabular} \\
\hline
Ye et al. \citep{ye2021} & \begin{tabular}[c]{@{}c@{}}Gene Ontology, KEGG, \\ STRING, MIPS\end{tabular} & \begin{tabular}[c]{@{}c@{}}Mashup algorithm, Deep Neural \\ Network, Extreme Gradient Boosting, \\ Logistic Regression, Naive Bayes\end{tabular} \\
\hline
Nunes et al. \citep{nunes2023} & \begin{tabular}[c]{@{}c@{}}Gene Ontology, Human \\ Phenotype Ontology, DisGeNET\end{tabular} & \begin{tabular}[c]{@{}c@{}}KG embedding methods (TransE, \\ HAKE, DistMult, RDF2Vec, OPA2Vec, \\ OWL2Vec), Random Forest, \\ Extreme Gradient Boosting\end{tabular} \\
\hline
Sousa et al. \citep{sousa2023explainable} & \begin{tabular}[c]{@{}c@{}}Human Phenotype Ontology, \\ DisGeNET \end{tabular} & \begin{tabular}[c]{@{}c@{}}KG embedding methods (RDF2Vec, \\ OWL2Vec*, TransE, TransH, DistMult), \\ Random Forest, Extreme Gradient \\ Boosting, Multi-layer Perceptron, SEEK \end{tabular} \\
\hline
Sousa et al. \citep{sousa2023negative} & \begin{tabular}[c]{@{}c@{}}Gene Ontology, \\ Human Phenotype Ontology, \\ Gene-disease Association dataset\end{tabular} & \begin{tabular}[c]{@{}c@{}}KG embedding methods (TransE, TransH, \\ TransR, ComplEx, DistMult, DeepWalk, \\ Node2Vec, Metapath2Vec, OWL2Vec*, \\ RDF2Vec, TrueWalks), Random Forest \end{tabular} \\ \hline

\end{tabular}%
}

\caption{Outline of the reviewed node-pair classification approaches for gene-disease association prediction.} \label{tab:ndapproaches}
\end{table}

\section{Materials and Methods}
\label{sec3}

\subsection{Experimental Design}
\label{subsec4}
\begin{figure}[t!]%
\centering
\includegraphics[width=\textwidth]{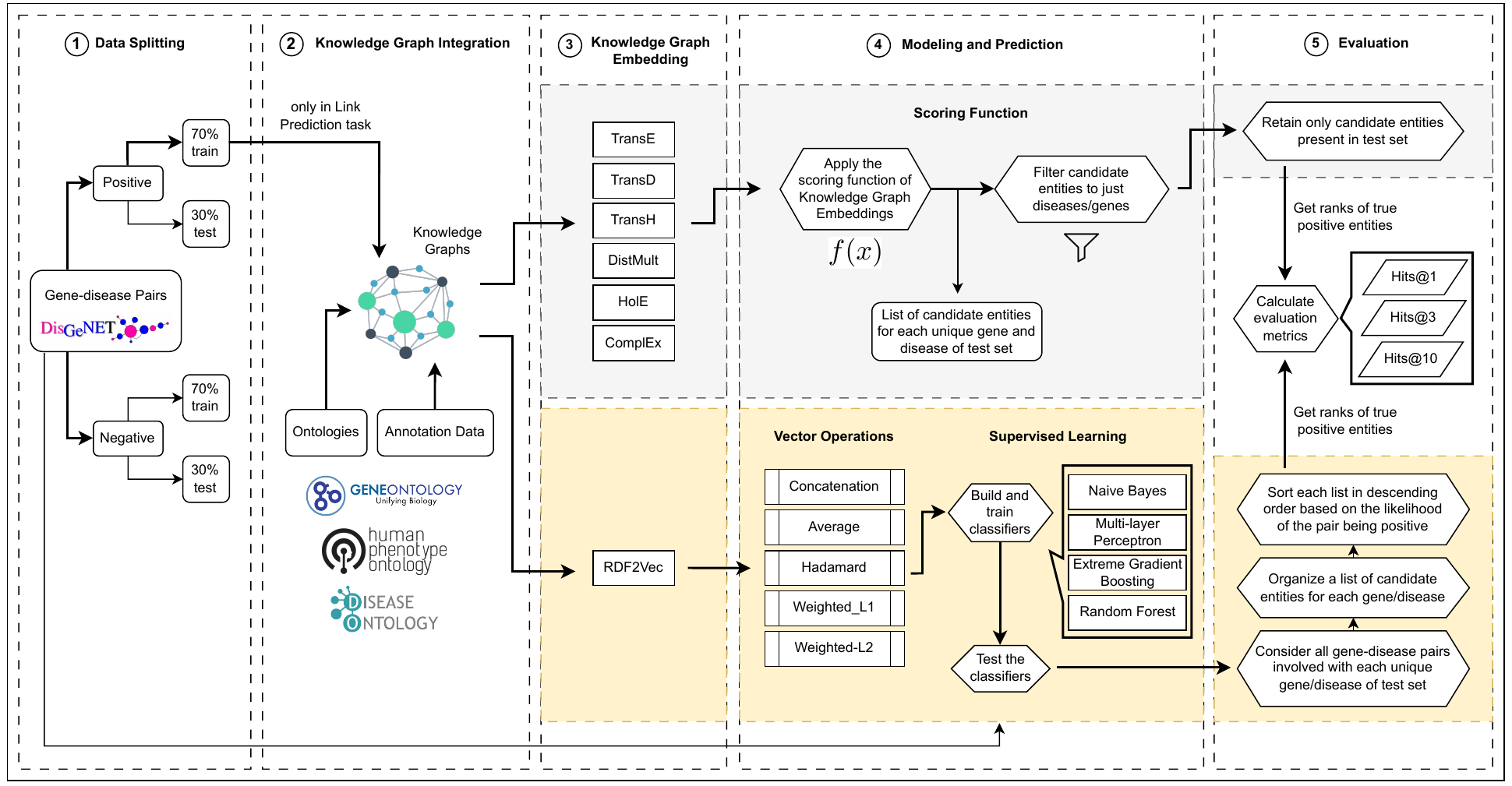}
\caption{Diagram of the comparison framework for link prediction and node-pair classification tasks. The framework consists of shared steps (data splitting, knowledge graph construction and evaluation) and task-specific steps: link prediction integrates positive training pairs into the knowledge graphs and applies the scoring function of knowledge graph embedding methods, whereas node-pair classification combines gene and disease embeddings, train supervised learning algorithms and test classifiers. Rectangles are color-coded: gray for link prediction-specific steps and yellow for node-pair classification-specific steps.} \label{workflow}
\end{figure}

Our comparison framework is based on different types of KGs with varying levels of richness both in the domains covered and in the types of links between ontologies. It also includes different KG embedding methods specifically tailored to link prediction and node-pair classification.

Figure \ref{workflow} outlines the proposed framework for comparing link prediction and node-pair classification tasks in gene-disease association prediction, which consists of five main steps. 

First, the collected target pairs --- positive and negative gene-disease pairs --- are split into 70\% for training and 30\% for testing. The same splits are used both in link prediction and node-pair classification tasks. Second, KGs with different characteristics are created (Section \ref{subsec5}) using ontologies, additional links between ontology classes, specifically logical definitions and ontology mappings, and positive training edges (only for the KGs used in the link prediction task).  Third, KG embedding methods are applied over the KGs to produce node and relation embeddings. The link prediction task applies translational distance models and semantic matching models, whereas the node-pair classification task uses a walk-based model. The choice behind using different KG embedding approaches for the two different tasks was motivated by the fact that Nunes et al. \citep{nunes2023} demonstrated that translational distance models achieved very low performance in a node-pair classification setting for gene-disease association prediction. The final embeddings were defined to be 200 features.

The fourth step corresponds to the prediction of gene-disease associations using either link prediction or node-pair classification. For link prediction (Section \ref{subsec6}), embeddings are given as input to the scoring functions of each KG embedding method for link prediction, which returns a list of candidate entities. Since link prediction methods predict links regardless of the type of node involved in them, the results were filtered only to contain predictions between diseases and genes. Link prediction is run both with genes as input entities and with diseases as input entities.

In node-pair classification (Section \ref{subsec7}), the embeddings of genes and diseases are combined, employing different aggregation approaches to produce representations of gene-disease pairs. These are then used as input features to train models using different supervised learning algorithms. The resulting models' performance is evaluated on the test set.

The last step corresponds to the evaluation. Typically, link prediction methods are evaluated using rank-based metrics, such as hits@k, mean rank, and mean reciprocal rank, whereas node-pair classification uses classification metrics, such as precision, recall, and f1-score. To support a more direct comparison between the two families of methods, we devised an approach to evaluate classification results using rank-based metrics (Section \ref{subsec8}).
The approach consists of unifying the results of the two families of methods into a single format: a list of candidate entities for each unique gene and for each unique disease of the test set, ensuring that the candidate entities belong exclusively to that set. In link prediction, the results are filtered to include only entities present in the test set.

In node-pair classification, a list of candidate entities is organized for each unique gene and disease, ordering the candidate entities by the probability of the gene-disease pair being positive.
With the results of both families of methods in the same format, the ranks of true positive entities are obtained, allowing the application of rank-based metrics.

\subsection{Building Knowledge Graphs}
\label{subsec5}

The proposed task comparison framework uses the Gene Ontology, the Human Phenotype Ontology, the Human Disease Ontology, logical definitions and ontology mappings between the first two ontologies, and gene-disease associations from DisGeNET to build different KGs. The inclusion of the Human Disease Ontology was motivated by the scarcity of approaches leveraging comprehensive disease ontologies, as among the reviewed gene-disease association approaches, none utilized a disease-specific ontology. Our study hypothesizes that including the Human Disease Ontology could improve method performance by offering a disease-specific and standardized vocabulary that complements the annotation and classification of diseases provided by the Human Phenotype Ontology.

The code and data used in the experiments are available at a \href{https://github.com/catarina-canastra/Benchmarking-Knowledge-Graph-Embeddings-for-Gene-Disease-Association-Prediction}{GitHub Repository}. \ref{app1} details the computational environment in which the experiments were carried out.

\subsubsection{Gene-Disease Associations}
\label{subsubsec3}

DisGeNET is one of the largest available collections of genes and variants involved in human diseases \citep{pinero2017}. It includes gene-disease associations extracted from multiple sources, including Uniprot \citep{uniprot2021}, OMIM \citep{hamosh2004}, and Orphanet \citep{pavan2017}, which are the same sources used to create some of the ontology annotations.

We have collected 16,378 gene-disease associations from Nunes et al. \citep{nunes2023}, with 50\% positive and the remaining negative pairs. The negative pairs were generated by randomly sampling the positive pairs. Random sampling assumes that negatives vastly outnumber positives, so negative instances are sampled more highly than positive ones \citep{olken1995}.

We performed a 70/30 split on the positive gene-disease pairs and the negative gene-disease pairs so that the training set (70\% of all pairs) had all types of entities and relationships, and the test set (30\% of all pairs) only had ``association" relationships between genes and diseases.

\subsubsection{Ontologies}
\label{subsubsec1}

The \textbf{Gene Ontology} (GO) describes the universe of concepts associated with gene product functions and how these functions relate to each other. A gene product function corresponds to the protein and non-coding RNA molecules produced by genes. The GO describes a gene product in terms of biological processes, molecular function, and cellular function \citep{go2000,geneontology}.

The \textbf{Human Phenotype Ontology} (HP) is a comprehensive biological and informatics resource for analyzing phenotypic abnormalities in human diseases. It organizes information into six independent sub-ontologies: phenotype abnormalities, clinical modifier, mode of inheritance, past medical history, blood group, and frequency of phenotypic abnormalities \citep{hpo2021}.

The \textbf{Human Disease Ontology} (DO) describes human diseases, their phenotypic characteristics, and related disease concepts within the medical vocabulary. It categorizes diseases into various types, including those caused by infectious agents, anatomical entities, cellular proliferation, mental health, metabolism, genetics, physical disorders, and syndromes \citep{do2022}.

We have downloaded the ontology files in OWL format from each ontology’s official website to ensure the use of the latest curated version of the ontology. Each of these files contains the hierarchy of concepts, relations between concepts, and logical constraints.

For each ontology, we generate a corresponding annotation file that maps real-world entities - such as Entrez Gene ID (for genes) or UMLS CUI (for diseases) - to all ontology terms, represented as IRIs (Internationalized Resource Identifiers). In these files, the first column contains the entity ID, while the second column lists all matching ontology terms from that specific ontology. This mapping serves as a bridge between different ontologies when the same concepts share the same identifier, and allows the inclusion of specific gene-disease associations from external datasets.

\subsubsection{Logical Definitions and Ontology Mappings}
\label{subsubsec2}

We have sourced logical definitions (LD) and ontology mappings (MAP) from Nunes et al. \citep{nunes2023}, who used AML-Compound (an AgreementMakerLigth ontology matching system variant) to retrieve relations between GO and HP ontology classes. They used an empirically determined threshold of 0.8 and found 494 MAPs, where 37 were identical to the existing LDs.

\subsubsection{Knowledge Graphs}
\label{subsubsec4}

The practical construction of KGs requires a selection of libraries, packages, and methodologies tailored to the intended objectives and domain-specific requirements. We used {\textit{RDFLib\footnote{\url{https://rdflib.readthedocs.io/en/stable/index.html/}}} (version 5.0.0) to create the KGs for input from KG embedding methods of the link prediction and node-pair classification tasks. \textit{RDFLib} is a \textit{Python} library well-suited for building KGs due to its robust support for creating, querying, and manipulating graph-based knowledge structures.

We mainly used \textit{.parse} and \textit{.add} methods to read and load OWL ontology files, add annotations, and include training gene-disease associations (in the case of link prediction) to the KGs. Finally, we used \textit{.serialize} method to save the KGs in XML format. RDF2Vec and the link prediction KG embedding methods take the annotation files in the format 'entity\_URL \textit{tab} list\_of\_annotations'. RDF2Vec also needs a file with all the entities appearing in the KGs, one entity per line with the full URL.

Table \ref{tab:statistics} presents the KGs created and summarizes relevant statistics regarding these KGs, namely: Classes (number of classes), Genes A. (number of annotations for genes), Diseases A. (number of annotations for diseases), LDs (number of LD) and MAPs (number of MAP) between GO and HP.

\begin{table}[h]
\renewcommand{\arraystretch}{1.5}
\centering
\begin{tabular}{lccccc}
\hline
\textbf{KGs} & \textbf{Classes} & \textbf{Gene A.} & \textbf{Disease A.} & \textbf{LDs} & \textbf{MAPs} \\ \hline
G+H & 294766 & 5901 & 1848 & - & - \\ \hline
G+H+L & 295118 & 5901 & 1848 & 350 & - \\ \hline
G+H+M & 295261 & 5901 & 1848 & - & 494 \\ \hline
G+H+L+M & 295580 & 5901 & 1848 & 350 & 494 \\ \hline
G+H+D & 324407 & 5901 & 8699 & - & - \\ \hline
G+H+D+L & 324759 & 5901 & 8699 & 350 & - \\ \hline
G+H+D+M & 324902 & 5901 & 8699 & - & 494 \\ \hline
G+H+D+L+M & 325221 & 5901 & 8699 & 350 & 494 \\ \hline
G+H*+D+L+M & 301532 & 2716 & 1848 & 350 & 494 \\ \hline
G+H*+D+L+M & 324752 & 2716 & 8699 & 350 & 494 \\ \hline
\end{tabular}%
\caption{Number of classes (Classes), gene annotations (Gene A.), disease annotations (Disease A.), logical definitions, and ontology mappings across individual knowledge graphs. Knowledge graphs details, the ontologies and links included in the knowledge graph.\textit{G: Gene Ontology.
H: Human Phenotype Ontology.
D: Disease Ontology.
L: Logical Definitions.
M: Ontology Mappings.
*: Knowledge graphs with HP annotations only for diseases.}} \label{tab:statistics}
\end{table}

The largest KG created - \textit{``G+H+D+L+M"}, consists of \textbf{8,881 genes}, \textbf{36,028 diseases}, and \textbf{5732 gene-disease pairs}. This KG represents the maximum number of genes and diseases involved in the experiments of this study. Figure \ref{semanticmodel} represents the maximum semantic model that KGs can have, where genes and diseases are depicted within circles, ontologies within squares, and LD and MAP are described among squares.

\begin{figure}[t]%
\centering
\includegraphics[width=\linewidth]{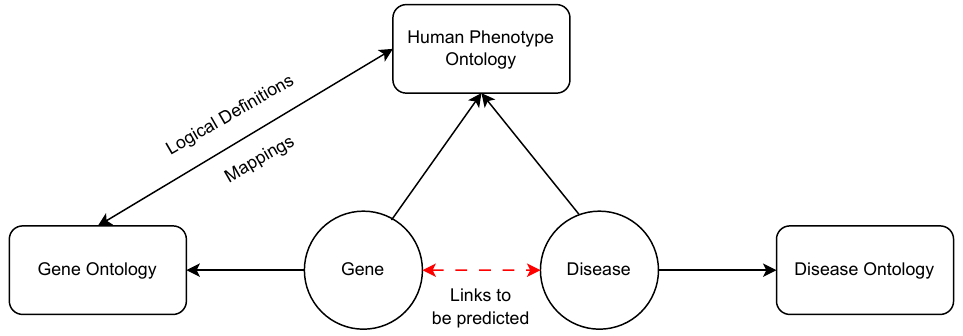}
\caption{Visual depiction illustrating interconnected relationships among genes, diseases, and ontologies, providing a deeper understanding of complex biological associations within a structured knowledge graph.}
\label{semanticmodel}
\end{figure}

\subsection{Link Prediction}
\label{subsec6}

The link prediction approach consists of three key steps after creating the KGs and before evaluating the methods:

\begin{enumerate}
    \item Employing translational distance models (TransE, TransD, and TransH) and semantic matching models (DisMult, HolE, and ComplEx) to obtain gene, disease, and association-relation embeddings;
    \item Passing the resulting embeddings into the scoring function of each KG embedding method to get a list of candidate entities for each input gene and input disease;
    \item Filtering the candidate entities to just diseases or genes, depending on whether it was an input gene or an input disease.
\end{enumerate}

The link prediction approach is based on Vilela et al. \citep{vilela2023}, where we applied three link prediction KG embedding methods (TransD, TransH, and HolE) in addition to the three that were used. These three additional methods were selected to ensure a more diverse set of KG embedding methods for link prediction task, which capture different perspectives on relational and structural patterns. These models offer complementary strengths: TransD and TransH address potential limitations in handling complex relations by introducing flexible projections, while HolE captures rich interactions between entities and relations through holographic representations.

We applied the translational distance models and semantic matching models using the \textit{OpenKE} library (powered by \textit{Tensorflow}). \textit{OpenKE} is an open-source library designed for knowledge embedding, providing tools and algorithms for representing KGs as dense vectors. To use this library, KGs must be represented using two types of files:

\begin{itemize}
    \item \textbf{Entity and Relationship Files}: A file listing all nodes (entities), another file listing all relations, and a third file only containing entities involved in the target relationship type - \textit{``association"}, which is limited to genes and diseases. Each entity and relationship in the KG must be assigned a unique identifier. These files have all components of the KG and corresponding identifiers, one per line;
    \item \textbf{Triple Files}: A file for all training triples and another file for all testing triples. These files start with the number of triples, and the following lines are in the format \textit{(e1,e2,rel)}, indicating a relation between entities \textit{e1} and \textit{e2}.
\end{itemize}

We used \textit{RDFLib} along with essential \textit{Python} libraries, such as {\textit{NumPy\footnote{\url{https://numpy.org/}}} (version 1.19.5), to create this required KG representation format. The algorithms were applied with the default parameters, as detailed in \ref{app2}, and conducted during 100 training epochs. 

Assuming we are predicting the genes associated with a disease, the next step of the link prediction approach consists of applying the scoring function to embeddings of the known disease and relationship, along with different potential entities. The scoring function calculates a score for each potential entity, indicating how well it fits or aligns with the specified disease and relationship. Higher scores suggest a stronger likelihood of those entities being true positives. We applied the scoring function of all KG embedding methods (Table \ref{tab:embeddinglp} for scoring functions \citep{dai2020}) for each unique gene and disease in the test set for all experiments (KGs).

\begin{table}[h]
\begin{center}
\begin{tabular}{ c c }
\toprule
 {\textbf{Model}} & {\textbf{Scoring Function $f_t (h, t)$}} \\
 \midrule
 TransE & $||h+r-t||_{l_1/l_2}$ \\[1pt]
 TransD & $||(r_p h^\parallel_p + I)h+r-(r_p t^\parallel_p + I)t||^2_2$ \\[1pt]
 TransH & $||(h-w^\parallel_r hw_r)+d_r-(t-w^\parallel_r tw_r)||^2_2$ \\[1pt]
 DistMult & $h^\parallel diag(r)t$ \\[1pt]
 HolE & $r^\parallel (h \star t)$ \\[1pt]
 ComplEx & $Re (h^\parallel diag(r) \overline{\text{t}})$ \\
\midrule
\end{tabular}
\end{center}

\caption{Scoring functions of the Knowledge Graph Embedding methods for link prediction.
\textit{h,t,r} represents embeddings of the head entity, tail entity, and relation, respectively.
$\|\cdot\|_{l_1/l_2}$ denotes the norm used to calculate distance. \(l_1\) norm is the Manhattan distance; \(l_2\) norm is the Euclidean distance.
\(r_p\) is the relation-specific projection matrix or vector.
\(h^\parallel_p, \, t^\parallel_p\) are projected embeddings of the head and tail entities based on the relation.
\(I\) is the identity matrix, indicating no transformation.
\(w^\parallel_r \) is the relation-specific projection vector.
\(d_r \) corresponds to the translation vector specific to the relation.
\(\text{diag}(r) \) indicates the diagonal matrix formed from the elements of the relation embedding vector \textit{r}.
\( \star \) is a convolution operation.
\( \overline{t} \) refers to the complex conjugate of the tail entity embedding \textit{t}.
 \(\text{Re}(\cdot) \) identifies the real part of the given complex number.
} 
\label{tab:embeddinglp}

\end{table}

The output is a list of all entities in the KG ordered in descending order of likelihood (or probability). This list is then filtered to only contain genes because it is the only type of entity that matches the ``association" relationship type.

\subsection{Node-Pair Classification}
\label{subsec7}

The approach for identifying and classifying gene-disease associations consists of four main steps:

\begin{enumerate}
    \item Employing a walk-based model - RDF2Vec, to obtain gene and disease embeddings;
    \item Combining the obtained embeddings for each gene-disease pair by five aggregation approaches: Concatenation, Average, Hadamard, Weighted-L1, and Weighted-L2;
    \item Training four supervised learning algorithms over the combined embeddings to predict non-represented gene-disease associations: Naive Bayes (NB), Multi-layer Perceptron (MLP), Extreme Gradient Boosting (XGB), and Random Forest (RF);
    \item Testing the classifiers with positive and negative gene-disease pairs not seen in the previous step;
\end{enumerate}

The node-pair classification approach is based on Nunes et al. \citep{nunes2023}. Regarding the KG embedding methods, we applied RDF2Vec because of its focus on generating embeddings through KG walks (paths) and Word2vec allows it to capture both the structural and semantic features of nodes well. Scoring-function-based algorithms, namely TransE and DistMult, are not directly applicable to node-pair classification because it focuses on relationship modelling. As for the aggregation approaches, we employed all aggregation approaches that were considered. Finally, we tested the performance of four classifiers, completing the original approach that only considered RF and XGB classifiers (both ensemble techniques) to ensure a more comprehensive evaluation with a broader spectrum of ML paradigms.

We used {\textit{PyRDF2Vec\footnote{\url{https://pyrdf2vec.readthedocs.io/en/latest/}}} to apply the RDF2Vec algorithm. \textit{PyRDF2Vec} is a \textit{Python} library that generates embeddings for entities by using a Skip-gram model, which learns vector representations by traversing the KG and considering the context in which entities occur. The algorithm was employed with the following parameters: 200-dimensional vectors; walks generated using the Weisfeiler-Lehman algorithm; maximum length of extracted walks 8; walks per entity 500; the corpora of walks were used to build a Skip-gram model with the default parameters for Word2Vec (\ref{app3} for default parameters).

After the KG embedding step, each gene-disease pair corresponds to two vectors, $f_i(g)$ and $f_i(d)$, associated with a gene $g$ and a disease $d$, respectively. To generate the pair representation $r(g,d)$ such that r: $V \times V \rightarrow \mathbb{R}^{d'}$ where $d'$ is the size of the pair $(g,d)$, we applied five different mathematical expressions (or operations) over the corresponding vectors to aggregate them, as summarized in Table \ref{tab:operators}.

\begin{table}[h]
\begin{center}
\begin{tabular}{ c c }
\toprule
 {\textbf{Operator}} & {\textbf{Definition}} \\
 \midrule
 Concatenation & $f_i (g) || g_i (d)$ \\[1pt]
 Average & $\frac{f_i (g) + g_i (d)}{2}$ \\[1pt]
 Hadamard & $f_i (g) \times g_i (d)$ \\[1pt]
 Weighted-L1 & $|f_i (g) - g_i (d)|$ \\[1pt]
 Weighted-L2 & $|f_i (g) - g_i (d)|^2$ \\
\midrule
\end{tabular}
\end{center}
\caption{Embedding aggregation operations.} \label{tab:operators}
\end{table}

Four classifiers were then built with the best set of parameters for each supervised learning algorithm according to Grid-Search exploration by Nunes et al. \citep{nunes2023}: NB, MLP, XGB and RF. \ref{app4} exhibits the parameter sets tested during the Grid-Search exploration and highlights the parameters used by the algorithms in the experiments of this study. Training pair representations were fed to the classifiers so that they learned to identify positive (true) and negative (false) gene-disease associations. Finally, non-represented gene-disease associations (testing pairs) were presented to the classifiers for categorizing these pairs. 

The output is a list with the probability of each test gene-disease pair being positive, negative, the classifier prediction, and the correct label. The probabilities are the output of the \textit{predict\_proba} method of \textit{Scikit-learn Python} library which returns the probability of each sample (pair) belonging to each class.

\subsection{Evaluation Metrics}
\label{subsec8}

In order to apply the same evaluation metrics to both link prediction and node-pair classification results, the results must address the same format and must involve only entities of the test set, as node-pair classification methods make predictions exclusively for entities within the test set. Node-pair classification methods do not have the ability to explore the structure of the KG and rank entities from the entire set of entities within the KG.

The link prediction results were filtered to include only entities of the test set. Concerning the node-pair classification results, all gene-disease pairs involved with each unique gene and each unique disease of the test set were considered. Then, a list of candidate entities was organized for each gene and disease, and the lists were sorted in descending order based on the likelihood (probability) of each pair being a validated (positive) gene-disease association.

With the results of both families of methods being in the same format and consisting solely on entities of the test set, a rank was assigned to each candidate entity, with the highest probability candidate receiving rank 1. Only the ranks of the entities truly associated with the target entity (gene or disease) were then collected. To assess the performance of link prediction and node-pair classification methods, we analyzed a modified version of the hits@k metric for the top 1, 3 and 10.

Hits@k is a metric used to evaluate the proportion of the correctly predicted entities ranked in the top \textit{k} among all entities of the same type. The idea is to measure the effectiveness of the method by considering the presence of candidate entities within the top $k$ positions of the candidates' list:

\begin{equation}
    \textit{Hits@k} = \frac{1}{n} \sum_{i=1}^{n} hits_i
\end{equation}

where $n$ would be the number of gene-disease pairs in the test set and $hits_i$ is a binary indicator that is 1 if the candidate entity for the $i$-th gene/disease is within the top $k$ positions and 0 otherwise \citep{rossi2021}.

We analyzed hits@k as the proportion of the correctly predicted entities ranked in the top $k$ among the total number of entities associated with the input entity (gene or disease). The objective is to evaluate the models' performance in a context that simulates real-world conditions.

Each false negative, i.e., gene-disease pair that was a true association in the test set but not found by the methods, was represented by a fixed and very low rank of 1000. The choice of 1000 is justified by the following considerations: the value serves as a penalty proxy, accounting for the fact that not identifying these entities reflects a limitation of the method; it is a sufficiently large value that assumes the model missed identifying many true associations \citep{kunegis2007}.

\section{Results and Discussion}
\label{sec4}

The present chapter will initially focus on the performance of methods in predicting diseases associated with input genes and then in predicting genes associated with input diseases. In node-pair classification, a method was considered to be the combination of the KG embedding method, the aggregation approach, and the supervised learning algorithm.

\subsection{Predicting Diseases associated with Genes}
Table \ref{tab:resultsGDhits1} depicts the hits@1 scores for the node-pair classification method, which consists of the RDF2Vec embedding method combined with the Hadamard operator and XGB algorithm, and the scores for the different link prediction methods across all KGs in predicting diseases associated with input genes.

\begin{table}
\centering
\resizebox{\textwidth}{!}{
\begin{threeparttable}
\begin{tabular}{l|c|ccccc|ccc}
\toprule
& \textbf{Node-Pair Classification} && \multicolumn{7}{c}{\textbf{Link Prediction}} \\
\cline{2-10}
& \textbf{Walk-based}  && \multicolumn{3}{c}{\textbf{Translational Distance}} && \multicolumn{3}{c}{\textbf{Semantic Matching}} \\
\cline{2-10}
\textbf{Knowledge Graphs} & \textbf{RDF2Vec} && \textbf{TransE} & \textbf{TransD} & \textbf{TransH} && \textbf{DistMult} & \textbf{HolE} & \textbf{ComplEx} \\
\cline{1-10}
G+H & 0.528 & & 0.619 & 0.552 & 0.640 && 0.638 & 0.666 & 0.634 \\ 
G+H+L & 0.532 & & 0.622 & 0.548 & 0.657  && 0.643 & 0.645 & 0.629 \\ 
G+H+M & 0.532 & & 0.624 & 0.543 & 0.629 && 0.634 & 0.657 & 0.618 \\ 
G+H+L+M & 0.529 & & 0.612 & 0.556 & 0.647 && 0.658 & 0.666 & 0.603 \\ 
G+H+D & 0.529 & & 0.647 & 0.549 & 0.658 && 0.647 & 0.650 & 0.629 \\ 
G+H+D+L & 0.530 & & 0.643 & 0.552 & 0.660 && 0.662 & \textbf{0.732} & 0.621 \\ 
G+H+D+M & 0.530 & & 0.660 & 0.558 & 0.654 && 0.658 & 0.670 & 0.630 \\ 
G+H+D+L+M & \textbf{0.539} & & 0.667 & 0.550 & 0.638 && 0.634 & 0.631 & 0.614 \\
G+H*+L+M & 0.462 & & \textbf{0.688} & \textbf{0.595} & \textbf{0.686} && 0.623 & 0.696 & \textbf{0.689} \\ 
G+H*+D+L+M & 0.473 & & 0.661 & 0.563 & 0.653 && \textbf{0.672} & 0.641 & 0.639 \\
\hline
\end{tabular}

\caption{Predictive performance for diseases associated with input genes. Assessment of hits@1 for the different methods across node-pair classification tasks using the Hadamard operator and Extreme Gradient Boosting algorithm and link prediction task over all experiments. The best hits@1 score for each method is highlighted in bold.} \label{tab:resultsGDhits1}
\end{threeparttable}
}
\end{table}

The scores in the node-pair classification task were regular across all KGs, with hits@1 between 0.462 and 0.539. The best hits@1 performance was achieved when using the largest KG created (\textit{``G+H+D+L+M"}), indicating the advantage of including a disease-specific ontology and additional links between ontology classes.

The performance of the link prediction methods was, in general, superior to that of the node-pair classification method. The best hits@1 (0.732) was observed for HolE utilizing the three ontologies and LDs between GO and HP. By contrast, TransD was the method with the lowest performance (their best performance was 0.595), being the closest to the node-pair classification method. TransE, TransD, TransH, and ComplEx models obtained their best performance using the same KG - \textit{``G+H*+L+M"}.

Table \ref{tab:resultsGDhits3} shows the hits@3 scores for node-pair classification (RDF2Vec + Hadamard + XGB) and link prediction methods in predicting diseases from input genes across all KGs. The node-pair classification task yielded stable scores across all KGs, with 80\% of true positive diseases in the top 1. The best hits@1 performance (0.845) was achieved by employing the KG that combines GO with HP and LDs between these ontologies.

\begin{table}
\centering
\resizebox{\textwidth}{!}{
\begin{threeparttable}
\begin{tabular}{l|c|ccccc|ccc}
\toprule
& \textbf{Node-Pair Classification} && \multicolumn{7}{c}{\textbf{Link Prediction}} \\
\cline{2-10}
& \textbf{Walk-based}  && \multicolumn{3}{c}{\textbf{Translational Distance}} && \multicolumn{3}{c}{\textbf{Semantic Matching}} \\
\cline{2-10}
\textbf{Knowledge Graphs} & \textbf{RDF2Vec} && \textbf{TransE} & \textbf{TransD} & \textbf{TransH} && \textbf{DistMult} & \textbf{HolE} & \textbf{ComplEx} \\
\cline{1-10}
G+H & 0.842 & & 0.938 & 0.875 & 0.937 && 0.941 & 0.944 & 0.929 \\ 
G+H+L & \textbf{0.845} & & 0.925 & 0.883 & 0.941  && 0.934 & 0.957 & 0.930 \\ 
G+H+M & 0.840 & & 0.933 & 0.880 & 0.930 && 0.927 & 0.936 & 0.915 \\ 
G+H+L+M & 0.837 & & 0.929 & 0.886 & 0.949 && 0.937 & 0.954 & 0.911 \\ 
G+H+D & 0.841 & & 0.941 & 0.880 & 0.941 && 0.930 & 0.946 & 0.924 \\ 
G+H+D+L & 0.840 & & 0.940 & 0.879 & \textbf{0.953} && \textbf{0.949} & \textbf{0.970} & 0.922 \\ 
G+H+D+M & 0.837 & & 0.946 & 0.877 & 0.950 && 0.939 & 0.949 & 0.919 \\ 
G+H+D+L+M & 0.842 & & 0.955 & 0.885 & 0.924 && 0.923 & 0.934 & 0.918 \\
G+H*+L+M & 0.810 & & \textbf{0.957} & \textbf{0.899} & 0.948 && 0.928 & 0.960 & 0.945 \\ 
G+H*+D+L+M & 0.805 & & 0.943 & 0.898 & 0.942 && 0.932 & 0.933 & \textbf{0.948} \\
\hline
\end{tabular}

\caption{Predictive performance for diseases associated with input genes. Assessment of hits@3 for the different methods across node-pair classification task using the Hadamard operator and Extreme Gradient Boosting algorithm and link prediction task over all experiments. The best hits@3 score for each method is highlighted in bold.} \label{tab:resultsGDhits3}
\end{threeparttable}
}
\end{table}

By comparison, the scores in the link prediction task were higher, with HolE being the best method across most experiments. This method achieved a hits@1 of 0.970, utilizing the \textit{``G+H+D+L"} graph. Using the same KG, TransH and DistMult obtained their best performance, achieving 0.953 and 0.949, respectively. TransE, TransD and ComplEx performed better when HP annotations for genes were not included (H*), being only ComplEx better with DO.

Table \ref{tab:resultsGDhits10} presents the hits@10 scores for node-pair classification (RDF2Vec + Hadamard + XGB) and link prediction methods across all KGs for disease prediction. 
The scores of the node-pair classification method across the KGs were consistently high, being all above 0.984. The best hits@10 score (0.991) was realized by utilizing the KG that combines the three ontologies and LDs between GO and HP.

\begin{table}
\centering
\resizebox{\textwidth}{!}{
\begin{threeparttable}
\begin{tabular}{l|c|ccccc|ccc}
\toprule
& \textbf{Node-Pair Classification} && \multicolumn{7}{c}{\textbf{Link Prediction}} \\
\cline{2-10}
& \textbf{Walk-based}  && \multicolumn{3}{c}{\textbf{Translational Distance}} && \multicolumn{3}{c}{\textbf{Semantic Matching}} \\
\cline{2-10}
\textbf{Knowledge Graphs} & \textbf{RDF2Vec} && \textbf{TransE} & \textbf{TransD} & \textbf{TransH} && \textbf{DistMult} & \textbf{HolE} & \textbf{ComplEx} \\
\cline{1-10}
G+H & 0.989 & & 1.000 & 1.000 & 1.000 && 1.000 & 1.000 & 1.000 \\ 
G+H+L & 0.989 & & 1.000 & \underline{0.999} & 1.000  && 1.000 & 1.000 & \underline{0.999} \\ 
G+H+M & 0.989 & & 1.000 & 1.000 & 1.000 && 1.000 & 1.000 & 1.000 \\ 
G+H+L+M & 0.989 & & 1.000 & \underline{0.999} & 1.000 && 1.000 & 1.000 & 1.000 \\ 
G+H+D & 0.987 & & 1.000 & 1.000 & 1.000 && 1.000 & \underline{0.998} & 1.000 \\ 
G+H+D+L & \textbf{0.991} & & 1.000 & 1.000 & 1.000 && 1.000 & 1.000 & 1.000 \\ 
G+H+D+M & 0.989 & & 1.000 & 1.000 & 1.000 && \underline{0.999} & 1.000 & 1.000 \\ 
G+H+D+L+M & 0.990 & & 1.000 & 1.000 & 1.000 && 1.000 & 1.000 & 1.000 \\
G+H*+L+M & 0.985 & & 1.000 & 1.000 & 1.000 && 1.000 & 1.000 & 1.000 \\ 
G+H*+D+L+M & 0.986 & & 1.000 & 1.000 & 1.000 && 1.000 & 1.000 & 1.000 \\
\hline
\end{tabular}

\caption{Predictive performance for diseases associated with input genes. Assessment of hits@10 for the different methods across node-pair classification task using the Hadamard operator and Extreme Gradient Boosting algorithm and link prediction task over all experiments. The best hits@10 score for the node-pair classification method is highlighted in bold and the lowest hits@10 score for each link prediction method is underlined.} \label{tab:resultsGDhits10}
\end{threeparttable}
}
\end{table}

The scores in the link prediction task were almost all excellent (1.000). The two best methods were TransE and TransH, achieving 1.000 in all experiments.
The models in which only 99\% of truly positive diseases were in the top 10 were TransD for \textit{``G+H+L"} and \textit{``G+H+L+M"} experiments, DistMult for \textit{``G+H+D+M"}, and ComplEx for \textit{``G+H+L"}. The HolE method scored a hits@10 of 0.998 when using the KG that combines the three ontologies and their corresponding annotations.

\subsection{Predicting Genes associated with Diseases}

Table \ref{tab:resultsDGhits1} exhibits the hits@1 scores for the node-pair classification method, which consists of the RDF2Vec embedding method combined with the Hadamard operator and XGB algorithm, and the scores for the different link prediction methods across all KGs in predicting genes associates with input diseases. \ref{app5} provide the hits@1, hits@3 and hits@10 scores for the other node-pair classification methods across all KGs in predicting diseases associated with input genes. Similarly, \ref{app6} offer supplementary findings regarding hits@1, hits@3 and hits@10 for the other node-pair classification methods across all KGs in predicting genes associated with input diseases.

\begin{table}
\centering
\resizebox{\textwidth}{!}{
\begin{threeparttable}
\begin{tabular}{l|c|ccccc|ccc}
\toprule
& \textbf{Node-Pair Classification} && \multicolumn{7}{c}{\textbf{Link Prediction}} \\
\cline{2-10}
& \textbf{Walk-based}  && \multicolumn{3}{c}{\textbf{Translational Distance}} && \multicolumn{3}{c}{\textbf{Semantic Matching}} \\
\cline{2-10}
\textbf{Knowledge Graphs} & \textbf{RDF2Vec} && \textbf{TransE} & \textbf{TransD} & \textbf{TransH} && \textbf{DistMult} & \textbf{HolE} & \textbf{ComplEx} \\
\cline{1-10}
G+H & 0.329 & & 0.669 & 0.604 & \textbf{0.674}  && 0.613 & 0.671 & 0.605 \\
G+H+L & 0.331 & & 0.689 & 0.607 & 0.618 && 0.611 & 0.653 & 0.613 \\
G+H+M & 0.331 & & 0.658 & 0.625 & 0.625 && 0.638 & 0.628 & 0.623 \\
G+H+L+M & \textbf{0.332} & & 0.624 & 0.603 & 0.612 &&  0.602 & \textbf{0.689} & 0.623 \\
G+H+D & \textbf{0.332} & & 0.579 & 0.633 & 0.633 && 0.634 & 0.632 & 0.604 \\
G+H+D+L & 0.330 & & \textbf{0.719} & \textbf{0.641} & 0.633 && \textbf{0.648} & 0.646 & 0.593 \\
G+H+D+M & 0.331 & & 0.623 & 0.612 & 0.634 && 0.631 & 0.640 & 0.631 \\
G+H+D+L+M & 0.328 & & 0.579 & 0.607 & 0.632 && 0.606 & 0.641 & 0.649 \\
G+H*+L+M & 0.276 & & 0.613 & 0.582 & 0.568 && 0.620 & 0.615 & \textbf{0.663} \\
G+H*+D+L+M & 0.280 & & 0.607 & 0.520 & 0.633 && 0.645 & 0.563 & 0.586 \\
\hline
\end{tabular}

\caption{Predictive performance for genes associated with input diseases. Assessment of hits@1 for the different methods across node-pair classification task using the Hadamard operator and Extreme Gradient Boosting algorithm and link prediction task over all experiments. The best hits@1 score for each method is highlighted in bold.} \label{tab:resultsDGhits1}
\end{threeparttable}
}
\end{table}

The scores of the node-pair classification method were low across all KGs, with hits@1 between 0.276 and 0.332. The best hits@1 performance was achieved when using both the KG that combines GO and HP with LDs and MAPs between these ontologies, and the KG that combines the three ontologies and their corresponding annotations.

By comparison, the performance of link prediction methods were regular across all KGs, and superior to that of the node-pair classification method. The best hits@1 was observed for TransE in the \textit{``G+H+D+L"} experiment, achieving an hits@1 of 0.719. TransD and DistMult also achieved their best performance using this KG, ranking 0.641 and 0.648, respectively.

Table \ref{tab:resultsDGhits3} shows the hits@3 scores for node-pair classification (RDF2Vec + Hadamard + XGB) and link prediction methods in predicting genes associated with input diseases across all KGs. The node-pair classification task yielded scores between 0.476 and 0.521, with the best method being the one that used KG consisting of GO and HP with both types of additional links between ontologies (LDs and MAPs).

\begin{table}
\centering
\resizebox{\textwidth}{!}{
\begin{threeparttable}
\begin{tabular}{l|c|ccccc|ccc}
\toprule
& \textbf{Node-Pair Classification} && \multicolumn{7}{c}{\textbf{Link Prediction}} \\
\cline{2-10}
& \textbf{Walk-based}  && \multicolumn{3}{c}{\textbf{Translational Distance}} && \multicolumn{3}{c}{\textbf{Semantic Matching}} \\
\cline{2-10}
\textbf{Knowledge Graphs} & \textbf{RDF2Vec} && \textbf{TransE} & \textbf{TransD} & \textbf{TransH} && \textbf{DistMult} & \textbf{HolE} & \textbf{ComplEx} \\
\cline{1-10}
G+H & 0.517 & & 0.912 & 0.872 & 0.919 && 0.883 & 0.906 & 0.880 \\ 
G+H+L & 0.517 & & 0.958 & 0.870 & 0.904  && 0.857 & 0.893 & 0.880 \\ 
G+H+M & 0.511 & & 0.910 & 0.900 & 0.844 && 0.890 & 0.874 & 0.878 \\ 
G+H+L+M & \textbf{0.521} & & 0.936 & 0.867 & 0.891 && 0.870 & 0.911 & 0.881 \\ 
G+H+D & 0.519 & & 0.850 & 0.884 & \textbf{0.938} && 0.882 & 0.896 & 0.879 \\ 
G+H+D+L & 0.514 & & \textbf{0.961} & \textbf{0.904} & 0.878 && \textbf{0.909} & 0.891 & 0.876 \\ 
G+H+D+M & 0.514 & & 0.849 & 0.868 & 0.927 && 0.906 & 0.885 & 0.888 \\ 
G+H+D+L+M & 0.514 & & 0.865 & 0.869 & 0.832 && 0.858 & 0.897 & 0.917 \\
G+H*+L+M & 0.476 & & 0.883 & 0.859 & 0.858 && 0.883 & \textbf{0.917} & 0.875 \\ 
G+H*+D+L+M & 0.477 & & 0.886 & 0.828 & 0.884 && 0.895 & 0.874 & \textbf{0.926} \\
\hline
\end{tabular}

\caption{Predictive performance for diseases associated with input genes. Assessment of hits@3 for the different methods across node-pair classification task using the Hadamard operator and Extreme Gradient Boosting algorithm and link prediction task over all experiments. The best hits@3 score for each method is highlighted in bold.} \label{tab:resultsDGhits3}
\end{threeparttable}
}
\end{table}

The scores in the link prediction task were good (over 0.800), with TransE being the best method, achieving a hits@3 of 0.961. This method achieved its best performance using the KG that combines the three ontologies and LDs between GO and HP. Using the same KG, TransD and DistMult obtained their best scores, reaching 0.904 and 0.909, respectively.

Regarding their best performance, TransH was the second-best method (0.938), utilizing the \textit{``G+H+D"} graph. HolE and ComplEx performed better when HP annotations for genes were not included (H*), being ComplEx better with DO. Nonetheless, the performance of all link prediction methods was generally similar to one another.

Table \ref{tab:resultsDGhits10} depicts the hits@10 scores for node-pair classification (RDF2Vec + Hadamard + XGB) and link prediction methods across all KGs for gene prediction. The scores of the node-pair classification method across the KGs were consistently acceptable (above 0.708). The best hits@10 score (0.735) was realized by utilizing the simplest KG (\textit{``G+H"}).

\begin{table}
\centering
\resizebox{\textwidth}{!}{
\begin{threeparttable}
\begin{tabular}{l|c|ccccc|ccc}
\toprule
& \textbf{Node-Pair Classification} && \multicolumn{7}{c}{\textbf{Link Prediction}} \\
\cline{2-10}
& \textbf{Walk-based}  && \multicolumn{3}{c}{\textbf{Translational Distance}} && \multicolumn{3}{c}{\textbf{Semantic Matching}} \\
\cline{2-10}
\textbf{Knowledge Graphs} & \textbf{RDF2Vec} && \textbf{TransE} & \textbf{TransD} & \textbf{TransH} && \textbf{DistMult} & \textbf{HolE} & \textbf{ComplEx} \\
\cline{1-10}
G+H & \textbf{0.735} & & \textbf{1.000} & 0.995 & \textbf{1.000} && 0.998 & \textbf{1.000} & 0.994 \\ 
G+H+L & 0.731 & & \textbf{1.000} & 0.998 & \textbf{1.000}  && 0.997 & \textbf{1.000} & 0.995 \\ 
G+H+M & 0.725 & & \textbf{1.000} & 0.993 & 0.977 && 0.996 & 0.992 & 0.998 \\ 
G+H+L+M & 0.724 & & \textbf{1.000} & 0.995 & \textbf{1.000} && 0.992 & \textbf{1.000} & 0.993 \\ 
G+H+D & 0.728 & & 0.992 & 0.995 & \textbf{1.000} && 0.996 & 0.991 & 0.994 \\ 
G+H+D+L & 0.728 & & \textbf{1.000} & \textbf{1.000} & \textbf{1.000} && \textbf{1.000} & \textbf{1.000} & \textbf{1.000} \\ 
G+H+D+M & 0.728 & & 0.993 & 0.994 & \textbf{1.000} && \textbf{1.000} & 0.992 & \textbf{1.000} \\ 
G+H+D+L+M & 0.727 & & 0.976 & \textbf{1.000} & 0.994 && \textbf{1.000} & \textbf{1.000} & 0.997 \\
G+H*+L+M & 0.709 & & 0.988 & 0.992 & \textbf{1.000} && \textbf{1.000} & \textbf{1.000} & \textbf{1.000} \\ 
G+H*+D+L+M & 0.719 & & \textbf{1.000} & 0.985 & 0.993 && \textbf{1.000} & 0.996 & 0.995 \\
\hline
\end{tabular}

\caption{Predictive performance for diseases associated with input genes. Assessment of hits@10 for the different methods across node-pair classification task using the Hadamard operator and Extreme Gradient Boosting algorithm and link prediction task over all experiments. The best hits@10 score for each method is highlighted in bold.} \label{tab:resultsDGhits10}
\end{threeparttable}
}
\end{table}

The performance of the link prediction methods was very high, achieving a hits@10 of 1.000 in some experiments, such as those that involved KGs with GO and HP and additional links between these ontologies when using TransE. The method that obtained the highest number of excellence scores (1,000) was TransH, while the method that least positioned all true positive genes in the top 10 was TransD.

Overall, the link prediction methods outperformed the node-pair classification method. The performance of the methods in predicting diseases associated with input genes, and predicting genes associated with input diseases, increases as more links between GO and HP are added. Adding a disease-specific ontology to the KGs does not significantly improve the performance of methods for predicting gene-disease associations.

The performance of the node-pair classification method decreased substantially when HP annotations for genes were not included (H*). In contrast, the performance of link prediction methods increases without these annotations: especially TransE, TransD and ComplEx in disease prediction; and ComplEx in gene prediction. HolE was generally the best link prediction method for disease prediction, and TransE was usually the best method in gene prediction.

\subsection{Discussion}

Link prediction methods universally demonstrated higher performance compared to the node-pair classification method. However, the node-pair classification method was shown to be more consistent across KGs with different characteristics, except when HP annotations for genes are not included (H*).

Hadamard (element-wise multiplication) outperformed the other aggregation approaches when combining gene and disease embeddings, as it may better capture complex feature interactions while maintaining a compact representation. By comparison, concatenation increases the feature space and may introduce redundant information, the average smooths embeddings but may lose important relational details, while weighted-L1 focuses on absolute differences but may not capture interactions effectively, and weighted-L2 emphasizes squared differences, highlighting larger discrepancies while reducing sensitivity to small variations.

Extreme Gradient Boosting outperformed the other supervised learning algorithms, and its superiority may lie in its optimization method that optimizes both bias and variance through gradient boosting, reducing overfitting while improving predictive accuracy. Unlike NB, which assumes feature independence, and MLP, which requires extensive hyperparameter tuning and computational sources, XGB handles feature correlations while leveraging regularization. Compared to RF, which builds independent trees, XGBoost sequentially improves weak learners, resulting in a potentially more refined and accurate model.

Among the link prediction methods, semantic matching models are better choices for predicting diseases associated with input genes. Specifically, HolE is more effective for disease prediction using a KG that combines GO, HP and DO with LDs between the first two ontologies. Conversely, translational distance models are better choices for predicting genes associated with input diseases.

One disadvantage of link prediction methods is that they rely on the complex structure of the KG to rank entities for completing a relationship. While this allows capturing complex patterns, it also introduces challenges in isolating associations between specific node types, such as genes and diseases. Unlike node-pair classification, link prediction methods do not focus on a subset of relationships but instead evaluate numerous potential triples spanning the KG, potentially diluting their precision for specific tasks. This also has implications at the level of evaluation, since node-pair classification methods make predictions for all positive pairs in the test set alongside an equal number of negative pairs, whereas link prediction methods rank candidate entities for completing the test triples by evaluating potential relationships across the entire KG.

Overall, predictive performance for disease-gene association was lower than that observed for gene-disease association.
Predicting genes associated with input diseases is more complex than predicting diseases associated with input genes due to several potential interrelated factors. Diseases are often polygenic, meaning they are influenced by variations in many different genes, each potentially contributing minor effects. This genetic heterogeneity hinders the identification of clear and consistent patterns that can reliably help predict which genes are associated with the disease. Additionally, the interactions between these genes and their regulatory pathways are complex and often context-dependent. By contrast, diseases associated with genes tend to be more predictable, as the presence of specific genetic variations or mutations can serve as stronger, more direct indicators of particular diseases \citep{cerrone2019,visscher2021}.

Generally, the predictive performance of all methods increases with a more connected KG. Comparing the same KG in the version without and in the version with the DO, most methods achieve higher accuracy in predicting diseases associated with input genes when leveraging DO. Conversely, some methods perform better for predicting genes associated with input diseases without DO, though it remains a valuable resource that significantly boosts predictions in many cases.

\section{Case Studies in Gene-Disease Association}
\label{sec5}

To validate the ability of the best node-pair classification method (combines RDF2Vec with Hadamard and XGB), and the ability of the link prediction methods to rank (priotitize) gene-disease associations, we considered two case studies: the first considering an input gene, where we want to predict its truly asssociated diseases; and the second consisting of an input disease, where the objective is to predict its associated genes. Our criteria for choosing both the example gene and disease was to select entities that in the node-pair classification task had approximately the same number of positive (true) and negative (false) gene-disease pairs in the test set. A method is considered good if it ranks the positive entities associated with the input entity.

The first case study concerns the Methylenetetrahydrofolate Reductase gene (MTHFR), identified by Entrez ID 4524 \citep{floris2020}, which is associated with 14 diseases in this study. This gene encodes an enzyme involved in folic acid metabolism and the remethylation of homocysteine to methionine \citep{floris2020}. Table \ref{tab:casestudy1} shows the rank of each disease for the different methods, using the overall best KG in the previous experiments, \textit{``G+H+D+L"}.

\begin{table}[h]
\centering
\renewcommand{\arraystretch}{1.2}
\resizebox{\textwidth}{!}{
\begin{tabular}{llc|c|ccc|ccc}
\toprule
&&& \textbf{Node-Pair Class} & \multicolumn{6}{c}{\textbf{Link Prediction}} \\
\cmidrule{4-10}
&&& \textbf{Walk-based} & \multicolumn{3}{c}{\textbf{Translational Distance}} & \multicolumn{3}{c}{\textbf{Semantic Matching}} \\
\midrule
\textbf{CUI} & \textbf{Disease} &  \textbf{\# genes} & \textbf{RDF2Vec} & \textbf{TransE} & \textbf{TransD} & \textbf{TransH} & \textbf{DistMult} & \textbf{HolE} & \textbf{ComplEx} \\
\midrule
C0004352 & Autism & 45 & 1 & 2 & 3 & 2 & 4 & 2 & 1 \\
C1856059 & Mthfr Deficiency & 1 & 2 & - & - & - & - & - & - \\
C0276496 & Familial Alzheimer's disease & 14 & 3 & 5 & 7 & - & 3 & - & 6 \\
C0005684 & Urinary bladder cancer & 14 & 4 & 7 & 1 & - & - & - & 7 \\
C0006142 & Breast cancer & 56 & 5 & 1 & 2 & 3 & 2 & 3 & 3 \\
C0038454 & Brain Attack & 9 & 6 & 6 & 4 & - & 6 & - & 5 \\
C0001973 & Alcohol dependence & 29 & 7 & 3 & 6 & 5 & 5 & 1 & 4 \\
C0080178 & Spina bifida & 3 & 8 & - & - & - & - & - & - \\
C0152426 & Craniorachischisis & 7 & 10 & 9 & - & 6 & - & - & - \\
C1858991 & Vanishing white matter disease & 3 & 11 & - & - & - & - & - & - \\
C0009402 & Colorectal cancer & 30 & 13 & 4 & 5 & 1 & 1 & - & 8 \\
C0003873 & Rheumatoid arthritis & 21 & 14 & 8 & 9 & 4 & - & - & 2 \\
C3160733 & Thrombophilia due to thrombin defect & 2 & 17 & - & - & - & - & - & - \\
C0009081 & Congenital clubfoot & 1 & 20 & - & - & - & - & - & - \\
\bottomrule
\end{tabular}
}
\caption{
Ranks assigned to each disease associated with the ``MTHFR" gene (Entrez ID 4524) by the different methods using the \textit{``G+H+D+L"} knowledge graph. RDF2Vec corresponds to the best configuration for the node-pair classification task using the Hadamard aggregation approach and Extreme Gradient Boosting algorithm. ``-" corresponds to a disease that was not ranked by the method. 
}\label{tab:casestudy1}
\end{table}

The node-pair classification method ranked all 14 diseases, as a characteristic of the node-pair classification methods is that they make predictions for all positive and negative gene-disease pairs in the test set (i.e., for all positive and negative diseases/genes associated with the input gene/disease). This method demonstrated high effectiveness in ranking diseases associated with the gene as it consistently prioritized truly associated diseases over non-associated ones. Notably, among the top 14 ranked diseases, only two known associated diseases were not successfully classified within this range.

Regarding the link prediction methods, TransE was able to rank nine out of 14 diseases, followed by TransD and ComplEx with eigth ranked diseases. HolE achieved the worst performance, ranking only three out of 14 diseases. Although the link prediction methods did not rank all truly associated diseases among all entities in the test set, they effectively prioritized associated diseases, placing them within the top 10 rankings.

The node-pair classification method and all link prediction methods ranked the diseases identified by CUI C0004352 (``Autism"), CUI C0006142 (``Breast cancer") and CUI0001973 (``Alcohol dependence"). Within these diseases, both the node-pair classification method and the ComplEx ranked ``Autism" as the top disease. Curiously, the diseases ranked at the top 1 are those with more than 10 associated genes (in this study). This suggests that diseases with a broader genetic basis may be easier to identify and prioritize. Converlesy, diseases with up to three asssociated genes, such ``Mthfr Deficiency" (1 gene) and ``Thrombophilia due to thrombin defect" (2 genes), were just ranked by the node-pair classification method.

The second case study addresses genes associated with the Dysmyelopoietic Syndrome (CUI C3463824), which is associated with 16 genes in this study. This disease is a haematological condition characterized by changes in the formation and maturation of blood cells in the bone marrow \citep{daver2022}. Table \ref{tab:casestudy2} presents all 16 genes identified by their Entrez ID and the rank predicted for each gene by the evaluated methods, using the overall best KG in the previous experiments, \textit{``G+H+D+L"}.

\begin{table}[h]
\centering
\renewcommand{\arraystretch}{1.2}
\resizebox{\textwidth}{!}{
\begin{tabular}{llc|c|ccc|ccc}
\toprule
&&& \textbf{Node-Pair Class} & \multicolumn{6}{c}{\textbf{Link Prediction}} \\
\cmidrule{4-10}
&&& \textbf{Walk-based} & \multicolumn{3}{c}{\textbf{Translational Distance}} & \multicolumn{3}{c}{\textbf{Semantic Matching}} \\
\midrule
\textbf{Gene ID} & \textbf{Gene Symbol} & \textbf{\# diseases} &\textbf{RDF2Vec} & \textbf{TransE} & \textbf{TransD} & \textbf{TransH} & \textbf{DistMult} & \textbf{HolE} & \textbf{ComplEx} \\
\midrule
4609 & TRIM5 & 9 & 1 & - & - & 2 & - & - & - \\
83990 & BRIP1 & 3 & 2 & - & 2 & - & - & - & 4 \\
54790 & TET2 & 2 & 3 & - & - & - & - & - & - \\
6427 & SRSF2 & 1 & 4 & - & - & - & 3 & - & - \\
29089 & PPP1R26 & 2 & 5 & - & - & - & - & - & 5 \\
7157 & MMP11 & 12 & 6 & - & 6 & 1 & - & 2 & - \\
672 & GLI3 & 2 & 7 & - & 1 & 3 & - & 3 & 1 \\
2176 & FANCC & 1 & 8 & 1 & 4 & - & - & - & - \\
5889 & RAD51C & 2 & 9 & - & 5 & - & - & - & 6 \\
4352 & ASAP2 & 2 & 10 & - & - & - & 4 & 5 & - \\
23451 & SF3B1 & 3 & 11 & - & - & - & - & - & -\\
7097 & TLR2 & 3 & 14 & - & - & - & - & - & - \\
54809 & SAMD9 & 1 & 15 & - & - & - & - & - & - \\
2623 & FBN1 & 1 & 21 & - & - & - & 1 & - & 3 \\
23092 & ARHGAP26 & 1 & 24 & - & - & - & - & - & - \\
5781 & PTPN11 & 2 & 30 & - & 3 & - & - & 1 & 2 \\
\bottomrule
\end{tabular}
}
\caption{
Ranks were assigned to each gene associated with the 
``Dysmyelopoietic Syndrome" (CUI C3463824) by the different methods using the \textit{``G+H+D+L"} knowledge graph. RDF2Vec corresponds to the best configuration for the node-pair classification task using the Hadamard aggregation approach and Extreme Gradient Boosting algorithm. ``-" corresponds to a gene that was not ranked by the method.
}\label{tab:casestudy2} 
\end{table}

The node-pair classification method classified all 16 genes, being the top gene the 
``TRIM5", identified by Entrez ID 4609. This method demonstrated high ability to prioritize truly associated genes over non-associated ones. Among the top 14, only two known associated genes were not successfully classified within this range.

Concerning the link prediction methods, both TransD and ComplEx were able to rank six out of 16 genes. The ``GLI3" gene was the one ranked by the largest number of methods, where TransD and ComplEx prioritize it as the top gene. TransE achieved the worst performance, ranking only one out of 16 genes. Nonetheless, the link prediction methods ranked associated genes within the top 6.

The node-pair classification method and TranD ranked the genes identified by Entrez ID 83990 (``BRIP1") and Entrez ID 7157 (``MMP11") in the same position. Genes with more associated diseases, such as ``TRIM5" (9 diseases) and ``MMP11" (12 diseases), received better ranks, which suggests that the methods consider this type of entities more relevant. In contrast, genes with fewer associated diseases, such as ``SAMD9" (1 disease) and ``ARHGAP26" (1 disease) obteined higher rankings, indicating lower priority in the ranking.

\section{Conclusion}
\label{sec6}

Discovering gene-disease links is an important area of research with applications to understand disease origin and develop new prevention, diagnosis, and therapy techniques. Computational approaches based on KGs and ML provide a robust framework for gene-disease association prediction, addressing the complexity and scale of biological data. In particular, KGs enriched with biomedical ontologies offer a standardized vocabulary for describing biological entities and processes. However, most works investigating ontologies for gene-disease association prediction use highly limited representations of diseases, focusing solely on their phenotypes.

The gene-disease association problem can be naturally framed as a link prediction task, as the core question revolves around predicting an edge (i.e., an association) between two nodes. Nonetheless, several works frame this problem as a node-pair classification task. This study proposes a new framework for comparing the performance of KG-based link prediction against node-pair classification tasks, analyses the performance of state-of-the-art link prediction and node-pair classification approaches, and compares various order-based formulations of gene-disease association prediction and how different approaches perform under each of them. This study also assesses the impact of KG semantic richness, focusing on an improved representation of diseases and additional links between ontologies.

Our novel framework is characterized by five distinct steps: (1) splitting the positive and negative gene-disease pairs into training and testing sets; (2) building KGs with different sets of ontologies and additional links between the classes of these; (3) applying KG embedding methods over the KGs to produce node and relation embeddings (4) \textit{in link prediction:} giving the embeddings to the scoring functions of KG embedding methods to obtain a list of candidate entities; \textit{in node-pair classification:} combining the embeddings of genes and diseases, using the resulting pair representations as input features to train supervised learning algorithms, evaluating the classifiers, and organizing the predictions for each input entity; and (5) picking the ranks of true positive genes/diseases and evaluating the performance of the methods using rank-based metrics.

The experimental results have shown that KGs with additional links between ontology classes support an improved performance of both node-pair classification and link prediction methods. The annotations for HP significantly enhance the performance of node-pair classification methods. By contrast, link prediction methods such as Complex, TransE, and TransH perform best on KGs that do not have these annotations. The experimental results also have shown that DO contributes to the overall prediction of gene-disease associations, but does not significantly enrich the KGs.

For predicting diseases associated with input genes, the best combination of ontologies, along with their annotations and additional links between ontology classes, in the node-pair classification task was found to be KGs that combine, at least, GO with HP and LDs between these ontologies. The two best KGs for disease prediction in the link prediction task were found to be \textit{``G+H+D+L"} and \textit{``G+H*+L+M"}. For predicting genes associated with input diseases, the best KG in the node-pair classification task was found to be \textit{``G+H+L+M"}, whereas the best KG in the link prediction task was found to be \textit{``G+H+D+L"}.

The distinction in results suggests that link prediction methods are better at using the semantic richness encoded in KGs, as the optimal composition of KGs entirely depends on the method. The link prediction task focuses on discovering potential links based on the structure and properties of the KG, possibly benefiting more from the multi-dimensional semantic information provided by combining GO, HP, DO, LDs, and MAPs. Among the link prediction methods, semantic matching models are better for predicting diseases associated with input genes, especially HolE, whereas translational distance models are better for gene prediction, particularly TransE.

Although link prediction methods perform better in predicting gene-disease associations, node-pair classification methods are designed to rank all true positives and negatives from the test set. In order to select the most appropriate task for a problem that can be naturally framed as a link prediction task, it is important to consider some methodological differences:

\begin{enumerate}
    \item Link prediction focuses on the structure and properties of the KG, while node-pair classification targets the attributes of individual nodes;
    \item Link prediction does not require the synthetic generation of negative examples, whereas node-pair classification requires generating or acquiring a balanced number of negative instances;
    \item KG embedding methods for link prediction are trained with positive pairs, which allows us to explore another aspect of semantic richness. Conversely, in node-pair classification, the pairs to be classified are processed only by a more traditional ML algorithm after obtaining the embeddings of the nodes;
    \item Whereas link prediction approaches typically involve a single algorithm -- KG embedding method, node-pair classification approaches use an embedding technique alongside a more traditional ML algorithm;
    \item The output of a link prediction method is a list of candidate entities for an input entity, limited to the number of entities of that type in the KG. In contrast, the output of a node-pair classification method consists of all positive and negative entities from the test set that are associated with the input entity.
\end{enumerate}

In summary, the advantages of framing a link prediction problem into a link prediction task over a node-pair classification task lie in its ability to leverage the rich semantic information encoded within KGs, uncover hidden relationships between entities of interest and promote more comprehensive predictive modelling in computational biology. This approach deepens our understanding of gene-disease associations and could also lead to valuable insights into disease mechanisms and the identification of potential therapeutic targets.

The present study cleared the way for further studies, developments, and optimizations. Regarding KG enrichment, possible paths could be the inclusion of a protein-protein interaction network in some KGs, relying upon the notion that protein interactions provide a functional context for molecular perturbations, including some disease mutations \citep{stanfield2017}; testing other gene and disease-specific ontologies, such as Sequence Ontology \citep{eilbeck2005}; exploring different databases (e.g., ClinVar \citep{landrum2016}); using subgraphs within KGs \citep{ji2024,zhang2024hcmg}; and employing clustering algorithms (e.g., Markov Clustering) to identify groups of related genes or diseases \citep{biswas2019}.

Concerning KG representation learning, possible paths could be exploring other KG embedding methods (e.g., TranSparse or KG2E) or refining the embeddings with the embeddings of other methods/algorithms. Another option could be to use models directly predicting gene-disease associations, such as graph neural networks \citep{he2021}. To fully realize the potential of the proposed framework, it is essential to explore and validate its application in diverse contexts. By extending its use to alternative domains, datasets, and approaches, researchers can refine its performance and contribute to a deeper understanding of its adaptability and effectiveness across varied scenarios.

\appendix
\section{Experimental Setup}
\label{app1}

The experiments were conducted on a server machine with a 12-core processor, 128GB RAM, and two NVIDIA Geforce RTX 2060 Super graphics cards, each with 8GB of VRAM. This server machine was essential, as its operating system offers more versatility in handling libraries and packages, and its graphics card meets the requirements of the \textit{OpenKE} library. Within the home, we created a \textit{Python} environment and installed: {\textit{grpcio\footnote{\url{https://pypi.org/project/grpcio/}}} (version 1.48.2); {\textit{HDF5\footnote{\url{https://docs.h5py.org/en/stable/index.html/}}} (version 3.1.0); {\textit{NumPy\footnote{\url{https://numpy.org/}}} (version 1.19.5); {\textit{Pandas\footnote{\url{https://pandas.pydata.org/}}} (version 1.1.5); \textit{RDFLib} (version 5.0.0); {\textit{Scikit-learn\footnote{\url{https://scikit-learn.org/stable/index.html/}}} (version 0.24.2); {\textit{Scipy\footnote{\url{https://pypi.org/project/scipy/}}} (version 1.5.4); and {\textit{TensorFlow\footnote{\url{https://www.tensorflow.org/?hl=pt/}}} (version 1.13.1). Installing all these libraries and packages guarantees the normal functioning of the experiments.

\section{Parameters of Knowledge Graph Embedding Methods for Link Prediction}
\label{app2}

\begin{table}[p]
\centering
\resizebox{\textwidth}{!}{
\begin{tabular}{lc}
\hline
\textbf{Algorithm} & \textbf{Parameters} \\
\hline
TransE & \begin{tabular}[c]{@{}l@{}} work\_threads = 8, nr\_batches = 100, alpha = 0.001, bern = 0, margin = 1.0,\\ entity\_negative\_rate = 1, relation\_negative\_rate = 0, optimization\_method = SGD \end{tabular} \\ \hline
TransD & \begin{tabular}[c]{@{}l@{}} work\_threads = 8, nr\_batches = 100, alpha = 1.0, bern = 1, margin = 4.0,\\ entity\_negative\_rate = 1, relation\_negative\_rate = 0, optimization\_method = SGD\end{tabular} \\ \hline
TransH & \begin{tabular}[c]{@{}l@{}} work\_threads = 8, nr\_batches = 100, alpha = 0.001, bern = 0, margin = 1.0,\\ entity\_negative\_rate = 1, relation\_negative\_rate = 0, optimization\_method = SGD\end{tabular} \\ \hline
DistMult & \begin{tabular}[c]{@{}l@{}} work\_threads = 8, nr\_batches = 100, alpha = 0.5, lambda = 0.05, bern = 1,\\ entity\_negative\_rate = 1, relation\_negative\_rate = 0, optimization\_method = Adagrad\end{tabular} \\ \hline
HolE & \begin{tabular}[c]{@{}l@{}} work\_threads = 8, nr\_batches = 100, alpha = 0.1, bern = 0, margin = 0.2,\\ entity\_negative\_rate = 1, relation\_negative\_rate = 0, optimization\_method = Adagrad\end{tabular} \\ \hline
ComplEx & \begin{tabular}[c]{@{}l@{}} work\_threads = 8, nr\_batches = 100, alpha = 0.5, lambda = 0.05, bern = 1,\\ entity\_negative\_rate = 1, relation\_negative\_rate = 0, optimization\_method = Adagrad\end{tabular} \\
\hline
\end{tabular}
}
\caption{Parameters of knowledge graph embedding methods for the link prediction task.} \label{tab:parametersLP}
\end{table}

\section{Parameters of Word2Vec of the Knowledge Graph Embedding Method for Node-Pair Classification}
\label{app3}

\begin{table}[p]
\centering
\resizebox{\columnwidth}{!}{%
\begin{tabular}{lc}
\hline
\textbf{Algorithm} & \textbf{Parameters} \\
\hline
\begin{tabular}[c]{@{}l@{}}RDF2Vec\\ (Word2Vec)\end{tabular} & \begin{tabular}[c]{@{}l@{}} sentences = None, corpus\_file = None, alpha = 0.025, window = 5,\\ min\_count = 5, max\_vocab\_size = None, sample = 0.001, seed = 1,\\ workers = 3, min\_alpha = 0.0001, sg = 0, hs = 0, negative = 5,\\ ns\_exponent = 0.75, hashfxn = 0, epochs = 5, null\_word = 0,\\ trim\_rule = None, sorted\_vocab = 1, batch\_words = 10000,\\ compute\_loss = False, callbacks = 0, comment = None,\\ max\_final\_vocab = None, shrink\_windows = True \end{tabular} \\
\hline
\end{tabular}
}
\caption{Default parameters of Word2Vec.} \label{tab:parametersW2V}
\end{table}

\section{Tested Parameter Sets during Grid-Search Exploration for the Supervised Learning Algorithms}
\label{app4}

\begin{table}[p]
\centering
\resizebox{\columnwidth}{!}{%
\begin{tabular}{lll}
\hline
\textbf{Algorithm} & \textbf{Parameter} & \textbf{Values} \\
\hline
RF & \begin{tabular}[c]{@{}l@{}}max\_depth:\\ n\_estimators:\end{tabular} & \begin{tabular}[c]{@{}l@{}}2, \textbf{4}, 6\\ 50, \textbf{100}, 200\end{tabular} \\ \hline
XGB & \begin{tabular}[c]{@{}l@{}}max\_depth:\\ n\_estimators:\\ learning\_rate:\end{tabular} & \begin{tabular}[c]{@{}l@{}}2, \textbf{4}, 6\\ 50, \textbf{100}, 200\\ \textbf{0.1}, 0.01, 0.001\end{tabular} \\ \hline
MLP & \begin{tabular}[c]{@{}l@{}}solver:\\ alpha:\\  hidden\_layer\_sizes:\\ random\_state:\end{tabular} & \begin{tabular}[c]{@{}l@{}}sgd, \textbf{adam}\\ \textbf{0.0001}, 0.05\\ \textbf{(10,10)}, (50,50), (100,100)\\ \textbf{1}, 5, 10\end{tabular} \\
\hline
\end{tabular}
}
\caption{Parameters sets for the supervised learning algorithms tested during the Grid-Search exploration by Nunes et al. \citep{nunes2023}. The values bolded represent the parameter values used by the algorithms in the experiments of the present study.} \label{tab:hyperparameters}
\end{table}

\section{Scores of hits@1, hits@3 and hits@10 for the Other Node-Pair Classification Methods in Predicting Diseases Associated with Input Genes}
\label{app5}

\begin{table}[p]
\centering
\resizebox{\columnwidth}{!}{%
\begin{tabular}{|l|c|ccc|ccc|ccc|ccc|}
\hline
\multirow{2}{*}{\bf \makecell{KG}} & \multirow{2}{*}{\bf \makecell{Operator}} & \multicolumn{12}{c|}{\bf \makecell{ Supervised Algorithm}} \\ \cline{3-14}
 &  & \multicolumn{3}{c}{\bf NB} & \multicolumn{3}{c}{\bf MLP} & \multicolumn{3}{c}{\bf XGB} & \multicolumn{3}{c|}{\bf RF} \\ \hline
 
\multirow{5}{*}{\textbf{G+H}} & \textbf{Concatenation} & 0.419 & 0.770 & 0.982 & 0.496 & 0.833 & 0.989 & 0.506 & 0.833 & 0.988 & 0.456 & 0.804 & 0.985 \\
 & \textbf{Average} & 0.490 & 0.814 & 0.981 & 0.514 & 0.838 & 0.989 & 0.515 & 0.830 & 0.989 & 0.491 & 0.816 & 0.987 \\
 & \textbf{Hadamard} & 0.490 & 0.811 & 0.979 & 0.353 & 0.840 & 0.991 & 0.528 & 0.842 & 0.989 & 0.506 & 0.826 & 0.987 \\
 & \textbf{Weighted-L1} & 0.523 & 0.835 & 0.988 & 0.502 & 0.826 & 0.987 & 0.519 & 0.836 & 0.990 & 0.507 & 0.826 & 0.989 \\
 & \textbf{Weighted-L2} & 0.525 & 0.834 & 0.988 & 0.502 & 0.826 & 0.987 & 0.519 & 0.836 & 0.990 & 0.510 & 0.829 & 0.987 \\ \hline
 
\multirow{5}{*}{\makecell{G+H+\\L}} & \textbf{Concatenation} & 0.421 & 0.768 & 0.982 & 0.489 & 0.836 & 0.990 & 0.505 & 0.827 & 0.987 & 0.451 & 0.805 & 0.985 \\
 & \textbf{Average} & 0.493 & 0.816 & 0.980 & 0.497 & 0.828 & 0.988 & 0.508 & 0.833 & 0.988 & 0.487 & 0.810 & 0.984 \\
 & \textbf{Hadamard} & 0.490 & 0.813 & 0.981 & 0.522 & 0.840 & 0.989 & 0.532 & 0.845 & 0.989 & 0.522 & 0.833 & 0.988 \\
 & \textbf{Weighted-L1} & 0.529 & 0.835 & 0.989 & 0.500 & 0.829 & 0.987 & 0.523 & 0.835 & 0.988 & 0.523 & 0.833 & 0.986 \\
 & \textbf{Weighted-L2} & 0.525 & 0.835 & 0.990 & 0.500 & 0.829 & 0.987 & 0.523 & 0.835 & 0.988 & 0.517 & 0.829 & 0.985 \\ \hline
 
\multirow{5}{*}{\makecell{G+H+\\M}} & \textbf{Concatenation} & 0.420 & 0.775 & 0.982 & 0.500 & 0.835 & 0.989 & 0.522 & 0.829 & 0.987 & 0.457 & 0.798 & 0.985 \\
 & \textbf{Average} & 0.496 & 0.810 & 0.980 & 0.512 & 0.840 & 0.989 & 0.519 & 0.834 & 0.988 & 0.500 & 0.825 & 0.988 \\
 & \textbf{Hadamard} & 0.493 & 0.807 & 0.980 & 0.520 & 0.840 & 0.989 & 0.532 & 0.840 & 0.989 & 0.522 & 0.828 & 0.988 \\
 & \textbf{Weighted-L1} & 0.533 & 0.828 & 0.989 & 0.495 & 0.818 & 0.987 & 0.527 & 0.836 & 0.989 & 0.512 & 0.829 & 0.989 \\
 & \textbf{Weighted-L2} & 0.527 & 0.825 & 0.986 & 0.495 & 0.818 & 0.987 & 0.527 & 0.836 & 0.989 & 0.515 & 0.832 & 0.987 \\ \hline
 
\multirow{5}{*}{\makecell{G+H+\\L+M}} & \textbf{Concatenation} & 0.424 & 0.769 & 0.980 & 0.497 & 0.841 & 0.987 & 0.500 & 0.826 & 0.987 & 0.460 & 0.799 & 0.985 \\
 & \textbf{Average} & 0.490 & 0.812 & 0.980 & 0.560 & 0.839 & 0.989 & 0.512 & 0.833 & 0.987 & 0.491 & 0.818 & 0.985 \\
 & \textbf{Hadamard} & 0.497 & 0.816 & 0.982 & 0.524 & 0.837 & 0.989 & 0.529 & 0.837 & 0.989 & 0.512 & 0.832 & 0.990 \\
 & \textbf{Weighted-L1} & 0.531 & 0.832 & 0.987 & 0.504 & 0.822 & 0.985 & 0.521 & 0.834 & 0.989 & 0.514 & 0.827 & 0.987 \\
 & \textbf{Weighted-L2} & 0.523 & 0.834 & 0.987 & 0.504 & 0.822 & 0.985 & 0.521 & 0.834 & 0.989 & 0.517 & 0.825 & 0.989 \\ \hline
 
\multirow{5}{*}{\makecell{G+H\\+D}} & \textbf{Concatenation} & 0.423 & 0.768 & 0.983 & 0.508 & 0.838 & 0.991 & 0.503 & 0.834 & 0.989 & 0.454 & 0.802 & 0.985 \\
 & \textbf{Average} & 0.507 & 0.821 & 0.983 & 0.508 & 0.836 & 0.991 & 0.508 & 0.829 & 0.989 & 0.497 & 0.821 & 0.987 \\
 & \textbf{Hadamard} & 0.501 & 0.815 & 0.979 & 0.525 & 0.834 & 0.987 & 0.529 & 0.841 & 0.987 & 0.507 & 0.829 & 0.989 \\
 & \textbf{Weighted-L1} & 0.523 & 0.836 & 0.988 & 0.498 & 0.821 & 0.987 & 0.523 & 0.837 & 0.989 & 0.514 & 0.827 & 0.988 \\
 & \textbf{Weighted-L2} & 0.516 & 0.832 & 0.987 & 0.498 & 0.821 & 0.987 & 0.523 & 0.837 & 0.989 & 0.514 & 0.836 & 0.989 \\ \hline
 
\multirow{5}{*}{\makecell{G+H+\\L+D}} & \textbf{Concatenation} & 0.420 & 0.775 & 0.983 & 0.496 & 0.832 & 0.989 & 0.512 & 0.829 & 0.989 & 0.462 & 0.807 & 0.986 \\
 & \textbf{Average} & 0.502 & 0.825 & 0.982 & 0.505 & 0.845 & 0.993 & 0.514 & 0.837 & 0.989 & 0.496 & 0.821 & 0.986 \\
 & \textbf{Hadamard} & 0.496 & 0.819 & 0.981 & 0.527 & 0.838 & 0.989 & 0.530 & 0.840 & 0.991 & 0.516 & 0.835 & 0.988 \\
 & \textbf{Weighted-L1} & 0.525 & 0.840 & 0.987 & 0.497 & 0.826 & 0.988 & 0.523 & 0.842 & 0.987 & 0.514 & 0.833 & 0.985 \\
 & \textbf{Weighted-L2} & 0.523 & 0.833 & 0.989 & 0.497 & 0.826 & 0.988 & 0.523 & 0.842 & 0.987 & 0.511 & 0.831 & 0.987 \\ \hline
 
\multirow{5}{*}{\makecell{G+H+\\D+M}} & \textbf{Concatenation} & 0.424 & 0.776 & 0.982 & 0.494 & 0.836 & 0.991 & 0.516 & 0.834 & 0.987 & 0.461 & 0.802 & 0.986 \\
 & \textbf{Average} & 0.580 & 0.820 & 0.981 & 0.506 & 0.841 & 0.988 & 0.514 & 0.834 & 0.989 & 0.497 & 0.815 & 0.986 \\
 & \textbf{Hadamard} & 0.502 & 0.821 & 0.979 & 0.524 & 0.840 & 0.987 & 0.530 & 0.837 & 0.989 & 0.512 & 0.825 & 0.989 \\
 & \textbf{Weighted-L1} & 0.521 & 0.835 & 0.988 & 0.510 & 0.829 & 0.987 & 0.522 & 0.836 & 0.989 & 0.520 & 0.834 & 0.985 \\
 & \textbf{Weighted-L2} & 0.519 & 0.832 & 0.987 & 0.510 & 0.829 & 0.987 & 0.522 & 0.836 & 0.989 & 0.509 & 0.831 & 0.987 \\ \hline
 
\multirow{5}{*}{\makecell{G+H+\\D+L+M}} & \textbf{Concatenation} & 0.423 & 0.772 & 0.982 & 0.515 & 0.837 & 0.989 & 0.515 & 0.836 & 0.989 & 0.453 & 0.798 & 0.987 \\
 & \textbf{Average} & 0.503 & 0.821 & 0.983 & 0.499 & 0.839 & 0.989 & 0.521 & 0.835 & 0.987 & 0.495 & 0.822 & 0.984 \\
 & \textbf{Hadamard} & 0.502 & 0.818 & 0.982 & 0.530 & 0.835 & 0.989 & 0.539 & 0.842 & 0.990 & 0.519 & 0.842 & 0.989 \\
 & \textbf{Weighted-L1} & 0.528 & 0.838 & 0.989 & 0.510 & 0.832 & 0.989 & 0.522 & 0.838 & 0.989 & 0.514 & 0.832 & 0.988 \\
 & \textbf{Weighted-L2} & 0.525 & 0.833 & 0.987 & 0.510 & 0.832 & 0.989 & 0.522 & 0.838 & 0.989 & 0.511 & 0.829 & 0.988 \\ \hline

 \multirow{5}{*}{\makecell{G+H*+\\L+M}} & \textbf{Concatenation} & 0.424 & 0.771 & 0.982 & 0.470 & 0.815 & 0.987 & 0.470 & 0.805 & 0.987 & 0.446 & 0.795 & 0.985 \\
 & \textbf{Average} & 0.433 & 0.784 & 0.984 & 0.474 & 0.817 & 0.989 & 0.472 & 0.814 & 0.985 & 0.449 & 0.795 & 0.985 \\
 & \textbf{Hadamard} & 0.438 & 0.789 & 0.981 & 0.451 & 0.810 & 0.989 & 0.462 & 0.810 & 0.985 & 0.453 & 0.797 & 0.984 \\
 & \textbf{Weighted-L1} & 0.440 & 0.794 & 0.983 & 0.457 & 0.796 & 0.986 & 0.442 & 0.791 & 0.986 & 0.432 & 0.789 & 0.983 \\
 & \textbf{Weighted-L2} & 0.443 & 0.789 & 0.984 & 0.457 & 0.796 & 0.986 & 0.442 & 0.791 & 0.986 & 0.437 & 0.786 & 0.895 \\ \hline

 \multirow{5}{*}{\makecell{G+H*+\\D+L+M}} & \textbf{Concatenation} & 0.418 & 0.772 & 0.982 & 0.476 & 0.820 & 0.986 & 0.471 & 0.812 & 0.986 & 0.446 & 0.796 & 0.982 \\
 & \textbf{Average} & 0.442 & 0.784 & 0.984 & 0.469 & 0.819 & 0.985 & 0.470 & 0.807 & 0.987 & 0.452 & 0.789 & 0.984 \\
 & \textbf{Hadamard} & 0.445 & 0.797 & 0.988 & 0.467 & 0.800 & 0.986 & 0.473 & 0.805 & 0.986 & 0.461 & 0.794 & 0.984 \\
 & \textbf{Weighted-L1} & 0.444 & 0.784 & 0.984 & 0.446 & 0.790 & 0.984 & 0.459 & 0.797 & 0.985 & 0.449 & 0.781 & 0.985 \\
 & \textbf{Weighted-L2} & 0.441 & 0.778 & 0.981 & 0.446 & 0.790 & 0.984 & 0.459 & 0.797 & 0.985 & 0.449 & 0.789 & 0.984 \\ \hline
\end{tabular}
}
\caption{Predictive performance for diseases associated with input genes. Assessment of hits@1, hits@3 and hits@10 for the different methods across node-pair classification task over all experiments, except the method that uses Hadamard and Extreme Gradient Boosting algorithm.} \label{geneDiseases}
\end{table}

% ----------------------------------------------------------------------------------

\section{Scores of hits@1, hits@3 and hits@10 for the Other Node-Pair Classification Methods in Predicting Genes Associated with Input Diseases}
\label{app6}

\begin{table}[p]
\centering
\resizebox{\columnwidth}{!}{%
\begin{tabular}{|l|c|ccc|ccc|ccc|ccc|}
\hline
\multirow{2}{*}{\bf \makecell{KG}} & \multirow{2}{*}{\bf \makecell{Operator}} & \multicolumn{12}{c|}{\bf \makecell{ Supervised Algorithm}} \\ \cline{3-14}
 &  & \multicolumn{3}{c}{\bf NB} & \multicolumn{3}{c}{\bf MLP} & \multicolumn{3}{c}{\bf XGB} & \multicolumn{3}{c|}{\bf RF} \\ \hline
 
\multirow{5}{*}{\textbf{G+H}} & \textbf{Concatenation} & 0.235 & 0.431 & 0.680 & 0.303 & 0.499 & 0.726 & 0.313 & 0.509 & 0.718 & 0.291 & 0.485 & 0.711 \\
 & \textbf{Average}& 0.327 & 0.507 & 0.721 & 0.306 & 0.504 & 0.729 & 0.316 & 0.506 & 0.729 & 0.305 & 0.491 & 0.712 \\
 & \textbf{Hadamard} & 0.329 & 0.507 & 0.723 & 0.329 & 0.514 & 0.729 & 0.329 & 0.517 & 0.735 & 0.320 & 0.503 & 0.724 \\
 & \textbf{Weighted-L1} & 0.332 & 0.515 & 0.722 & 0.313 & 0.498 & 0.713 & 0.322 & 0.508 & 0.725 & 0.321 & 0.499 & 0.718 \\
 & \textbf{Weighted-L2} & 0.332 & 0.511 & 0.719 & 0.313 & 0.498 & 0.713 & 0.322 & 0.508 & 0.725 & 0.322 & 0.503 & 0.720 \\ \hline
 
\multirow{5}{*}{\makecell{G+H+\\L}} & \textbf{Concatenation} & 0.239 & 0.433 & 0.684 & 0.300 & 0.492 & 0.723 & 0.312 & 0.504 & 0.723 & 0.293 & 0.485 & 0.718 \\
 & \textbf{Average}& 0.329 & 0.507 & 0.721 & 0.310 & 0.502 & 0.723 & 0.316 & 0.504 & 0.730 & 0.310 & 0.494 & 0.721 \\
 & \textbf{Hadamard} & 0.326 & 0.505 & 0.721 & 0.327 & 0.511 & 0.730 & 0.331 & 0.517 & 0.731 & 0.329 & 0.510 & 0.726 \\
 & \textbf{Weighted-L1} & 0.334 & 0.516 & 0.723 & 0.310 & 0.493 & 0.715 & 0.325 & 0.512 & 0.729 & 0.323 & 0.507 & 0.722 \\
 & \textbf{Weighted-L2} & 0.332 & 0.515 & 0.721 & 0.310 & 0.493 & 0.715 & 0.325 & 0.512 & 0.729 & 0.322 & 0.509 & 0.722 \\ \hline
 
\multirow{5}{*}{\makecell{G+H+\\M}} & \textbf{Concatenation} & 0.242 & 0.434 & 0.982 & 0.305 & 0.497 & 0.728 & 0.309 & 0.505 & 0.724 & 0.289 & 0.488 & 0.716 \\
 & \textbf{Average}& 0.326 & 0.502 & 0.720 & 0.305 & 0.501 & 0.728 & 0.322 & 0.508 & 0.732 & 0.314 & 0.498 & 0.720 \\
 & \textbf{Hadamard} & 0.330 & 0.500 & 0.716 & 0.328 & 0.515 & 0.728 & 0.331 & 0.511 & 0.725 & 0.324 & 0.504 & 0.719 \\
 & \textbf{Weighted-L1} & 0.332 & 0.512 & 0.723 & 0.306 & 0.489 & 0.718 & 0.325 & 0.508 & 0.719 & 0.322 & 0.500 & 0.712 \\
 & \textbf{Weighted-L2} & 0.332 & 0.508 & 0.722 & 0.306 & 0.489 & 0.718 & 0.325 & 0.508 & 0.719 & 0.319 & 0.504 & 0.711 \\ \hline
 
\multirow{5}{*}{\makecell{G+H+\\L+M}} & \textbf{Concatenation} & 0.246 & 0.436 & 0.676 & 0.300 & 0.501 & 0.726 & 0.310 & 0.506 & 0.728 & 0.290 & 0.487 & 0.711 \\
 & \textbf{Average}& 0.328 & 0.504 & 0.722 & 0.310 & 0.506 & 0.727 & 0.318 & 0.507 & 0.729 & 0.305 & 0.495 & 0.713 \\
 & \textbf{Hadamard} & 0.326 & 0.507 & 0.718 & 0.329 & 0.514 & 0.725 & 0.332 & 0.521 & 0.724 & 0.321 & 0.502 & 0.717 \\
 & \textbf{Weighted-L1} & 0.334 & 0.509 & 0.718 & 0.314 & 0.500 & 0.713 & 0.329 & 0.513 & 0.724 & 0.325 & 0.505 & 0.723 \\
 & \textbf{Weighted-L2} & 0.332 & 0.508 & 0.722 & 0.314 & 0.500 & 0.713 & 0.329 & 0.513 & 0.724 & 0.326 & 0.504 & 0.716 \\ \hline
 
\multirow{5}{*}{\makecell{G+H\\+D}} & \textbf{Concatenation} & 0.240 & 0.431 & 0.678 & 0.308 & 0.495 & 0.724 & 0.319 & 0.510 & 0.726 & 0.295 & 0.492 & 0.712 \\
 & \textbf{Average}& 0.331 & 0.510 & 0.722 & 0.305 & 0.492 & 0.722 & 0.317 & 0.507 & 0.724 & 0.315 & 0.498 & 0.711 \\
 & \textbf{Hadamard} & 0.325 & 0.504 & 0.718 & 0.321 & 0.509 & 0.733 & 0.332 & 0.519 & 0.728 & 0.327 & 0.511 & 0.722 \\
 & \textbf{Weighted-L1} & 0.334 & 0.522 & 0.721 & 0.316 & 0.490 & 0.714 & 0.332 & 0.516 & 0.726 & 0.326 & 0.505 & 0.718 \\
 & \textbf{Weighted-L2} & 0.333 & 0.518 & 0.725 & 0.316 & 0.490 & 0.714 & 0.332 & 0.516 & 0.726 & 0.331 & 0.510 & 0.722 \\ \hline
 
\multirow{5}{*}{\makecell{G+H+\\L+D}} & \textbf{Concatenation} & 0.239 & 0.431 & 0.683 & 0.300 & 0.492 & 0.726 & 0.312 & 0.505 & 0.722 & 0.294 & 0.481 & 0.717 \\
 & \textbf{Average}& 0.328 & 0.505 & 0.718 & 0.310 & 0.503 & 0.727 & 0.312 & 0.502 & 0.722 & 0.315 & 0.493 & 0.711 \\
 & \textbf{Hadamard} & 0.330 & 0.498 & 0.716 & 0.328 & 0.508 & 0.728 & 0.330 & 0.514 & 0.728 & 0.326 & 0.508 & 0.523 \\
 & \textbf{Weighted-L1} & 0.334 & 0.513 & 0.722 & 0.314 & 0.495 & 0.714 & 0.328 & 0.509 & 0.723 & 0.327 & 0.501 & 0.717 \\
 & \textbf{Weighted-L2} & 0.330 & 0.508 & 0.721 & 0.314 & 0.495 & 0.714 & 0.328 & 0.509 & 0.723 & 0.323 & 0.501 & 0.711 \\ \hline
 
\multirow{5}{*}{\makecell{G+H+\\D+M}} & \textbf{Concatenation} & 0.242 & 0.436 & 0.680 & 0.301 & 0.498 & 0.725 & 0.320 & 0.507 & 0.723 & 0.289 & 0.484 & 0.712 \\
 & \textbf{Average}& 0.330 & 0.503 & 0.722 & 0.306 & 0.501 & 0.728 & 0.320 & 0.504 & 0.725 & 0.306 & 0.493 & 0.716 \\
 & \textbf{Hadamard} & 0.331 & 0.503 & 0.721 & 0.322 & 0.505 & 0.728 & 0.331 & 0.514 & 0.728 & 0.325 & 0.513 & 0.722 \\
 & \textbf{Weighted-L1} & 0.338 & 0.518 & 0.720 & 0.311 & 0.493 & 0.711 & 0.330 & 0.512 & 0.722 & 0.329 & 0.508 & 0.722 \\
 & \textbf{Weighted-L2} & 0.335 & 0.518 & 0.719 & 0.311 & 0.493 & 0.711 & 0.330 & 0.512 & 0.722 & 0.328 & 0.507 & 0.721 \\ \hline
 
\multirow{5}{*}{\makecell{G+H+\\D+L+M}} & \textbf{Concatenation} & 0.243 & 0.436 & 0.684 & 0.307 & 0.502 & 0.720 & 0.311 & 0.503 & 0.726 & 0.287 & 0.479 & 0.717 \\
 & \textbf{Average}& 0.330 & 0.510 & 0.725 & 0.303 & 0.498 & 0.728 & 0.322 & 0.507 & 0.723 & 0.308 & 0.497 & 0.718 \\
 & \textbf{Hadamard} & 0.328 & 0.506 & 0.721 & 0.325 & 0.508 & 0.731 & 0.328 & 0.514 & 0.727 & 0.323 & 0.507 & 0.722 \\
 & \textbf{Weighted-L1} & 0.334 & 0.516 & 0.721 & 0.307 & 0.493 & 0.718 & 0.328 & 0.515 & 0.716 & 0.326 & 0.509 & 0.716 \\
 & \textbf{Weighted-L2} & 0.333 & 0.516 & 0.719 & 0.307 & 0.493 & 0.718 & 0.328 & 0.515 & 0.716 & 0.325 & 0.508 & 0.714 \\ \hline

 \multirow{5}{*}{\makecell{G+H*+\\L+M}} & \textbf{Concatenation} & 0.243 & 0.436 & 0.686 & 0.285 & 0.481 & 0.720 & 0.280 & 0.481 & 0.714 & 0.271 & 0.470 & 0.710 \\
 & \textbf{Average}& 0.269 & 0.455 & 0.698 & 0.283 & 0.485 & 0.718 & 0.282 & 0.478 & 0.716 & 0.270 & 0.460 & 0.701 \\
 & \textbf{Hadamard} & 0.263 & 0.449 & 0.704 & 0.278 & 0.470 & 0.712 & 0.276 & 0.476 & 0.709 & 0.279 & 0.466 & 0.713 \\
 & \textbf{Weighted-L1} & 0.277 & 0.468 & 0.706 & 0.274 & 0.467 & 0.697 & 0.266 & 0.458 & 0.704 & 0.272 & 0.463 & 0.705 \\
 & \textbf{Weighted-L2} & 0.269 & 0.462 & 0.701 & 0.274 & 0.467 & 0.697 & 0.266 & 0.458 & 0.704 & 0.265 & 0.459 & 0.705 \\ \hline

 \multirow{5}{*}{\makecell{G+H*+\\D+L+M}} & \textbf{Concatenation} & 0.245 & 0.436 & 0.678 & 0.293 & 0.482 & 0.721 & 0.287 & 0.478 & 0.704 & 0.274 & 0.466 & 0.701 \\
 & \textbf{Average}& 0.268 & 0.454 & 0.700 & 0.279 & 0.486 & 0.720 & 0.279 & 0.408 & 0.711 & 0.274 & 0.464 & 0.700 \\
 & \textbf{Hadamard} & 0.267 & 0.456 & 0.702 & 0.275 & 0.475 & 0.717 & 0.280 & 0.477 & 0.719 & 0.274 & 0.470 & 0.710 \\
 & \textbf{Weighted-L1} & 0.269 & 0.458 & 0.701 & 0.263 & 0.458 & 0.701 & 0.282 & 0.465 & 0.707 & 0.270 & 0.459 & 0.701 \\
 & \textbf{Weighted-L2} & 0.266 & 0.456 & 0.695 & 0.263 & 0.458 & 0.701 & 0282 & 0.465 & 0.707 & 0.268 & 0.461 & 0.702 \\ \hline
\end{tabular}
}
\caption{Predictive performance for genes associated with input diseases. Assessment of hits@1, hits@3 and hits@10 for the different methods across node-pair classification task over all experiments, except the method that uses Hadamard and Multi-layer Perceptron algorithm.} \label{diseaseGenes}
\end{table}

\section*{Acknowledgements}
\label{sec7}

This work was supported by FCT through the LASIGE Research Unit, ref. UID/00408/2025, partially supported by the KATY project which has received funding from the EU Horizon 2020 research and innovation program under grant agreement No 101017453, and also partially supported by project 41, HfPT: Health from Portugal, funded by the Portuguese Plano de Recuperação e Resiliência.

%%       \citep{<label>} ==> [1]

%%

%Example citation, See \citep{lamport94}.


%\bibliographystyle{elsarticle-num} 
%\bibliography{references}



%\begin{thebibliography}{00}

%% For numbered reference style

%\end{thebibliography}
\end{document}